\documentclass{fundam}

\publyear{2021}
\papernumber{2060}
\volume{182}
\issue{1}

\usepackage{url} 
\usepackage[ruled,lined]{algorithm2e}
\usepackage{graphicx}
\usepackage{tikz}
\usetikzlibrary{automata, positioning, arrows}
\usepackage{graphicx}
\usepackage{amsmath,amsfonts}
\usepackage{booktabs}
\usepackage{comment}
\usepackage{subcaption}
\usepackage{algpseudocode}
\begin{document}

\title{IV-GNN : Interval Valued Data Handling Using Graph Neural Network}

\author{Sucheta Dawn\\
Machine Intelligence Unit, Indian Statistical Institute, Kolkata, India\\
$suchetad\_r@isical.ac.in$
\and Sanghamitra Bandyopadhyay\\
Machine Intelligence Unit, Indian Statistical Institute, Kolkata, India \\
$sanghami@isical.ac.in$} 

\maketitle

\runninghead{S. Dawn, S. Bandyopadhyay }{IV-GNN : Interval Valued Data Handling Using Graph Neural Network}

\begin{abstract}
  Graph Neural Network (GNN) is a powerful tool to perform standard machine learning on graphs. To have a Euclidean representation of every node in the Non-Euclidean graph-like data, GNN follows neighbourhood aggregation and combination of information recursively along the edges of the graph. Despite having many GNN variants in the literature, no model can deal with graphs having nodes with interval-valued features. 
This article proposes an Interval-Valued Graph Neural Network, a novel GNN model where, for the first time, we relax the restriction of the feature space being countable. Our model is much more general than existing models as any countable set is always a subset of the universal set $\mathbb{R}^n$, which is uncountable. Here, to deal with interval-valued feature vectors, we propose a new aggregation scheme of intervals and show its expressive power to capture different interval structures. We validate our theoretical findings about our model for graph classification tasks by comparing its performance with those of the state-of-the-art models on several benchmark network and synthetic datasets. 
\end{abstract}


\section{Introduction}

\begin{figure}[htbp]
\centerline{\includegraphics[width=0.5\textwidth]{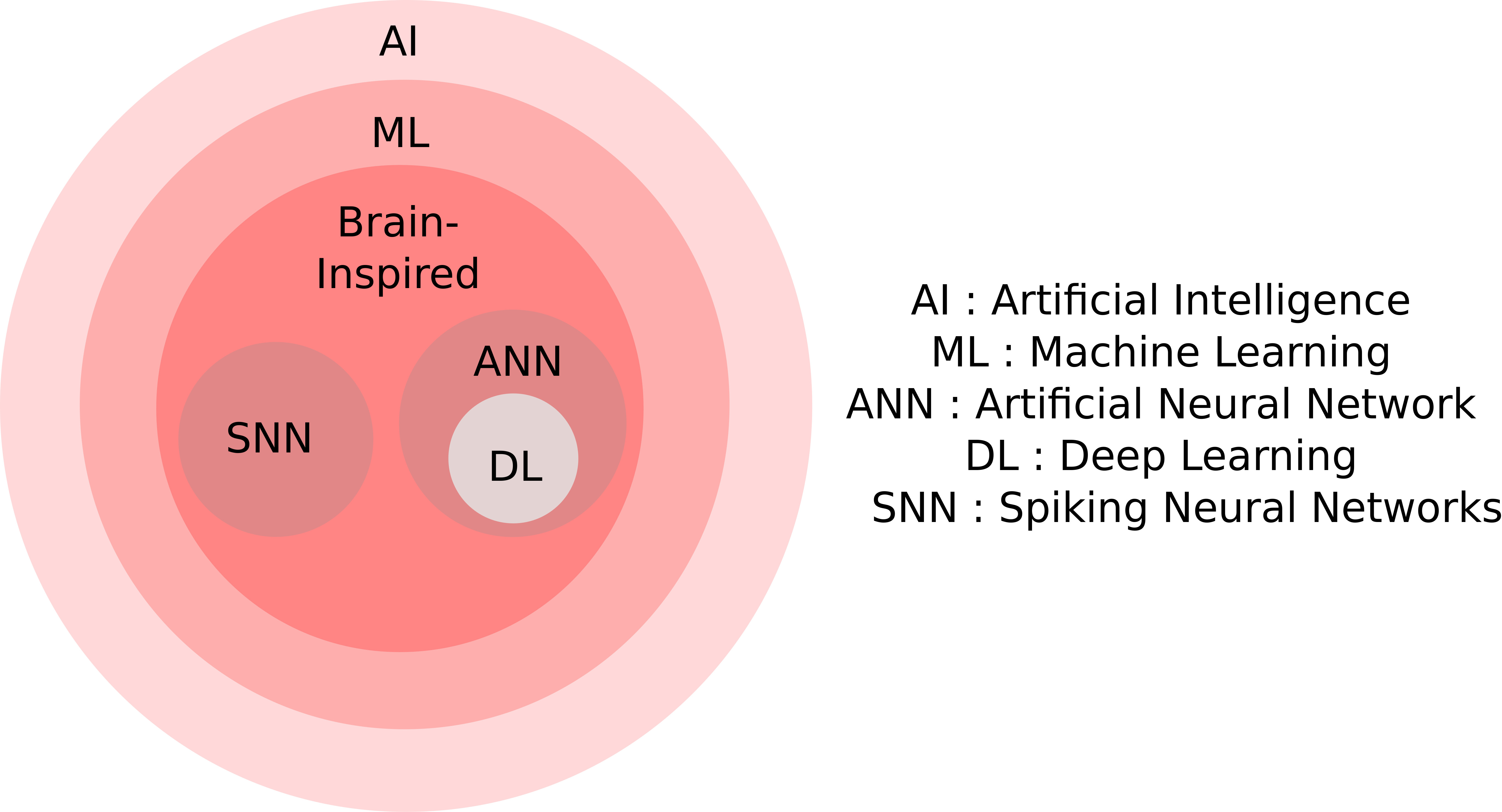}}
\caption{The taxonomy of AI \cite{alom2019state}. }
\label{figure1}
\end{figure}

Artificial Neural Network (ANN) is a branch of the general Machine Learning literature, which is inspired by biological neural networks (Fig.\ref{figure1}) \cite{alom2019state}. It has proven to work well in several Machine Learning related tasks such as Recognition, Classification, Prediction, and so on \cite{shafiee2016stochasticnet}. It has outperformed traditional Machine Learning techniques while solving problems in various fields like Image Processing, Computer Vision, Speech Recognition, Machine Translation, Bioinformatics, Natural Language Processing, and many others \cite{alom2019state}. Traditional Neural Network deals with those data which belong to the Euclidean space \cite{scarselli2008graph}. However, most of the data from the real-life scenario do not have a Euclidean structure behind them and can be represented better as Non-Euclidean data such as graphs and manifolds. For instance, in the e-commerce system \cite{ying2018graph}, the interactions between users and products can be represented as graphs. Additionally, bio-active molecules and their bio-activity \cite{defferrard2016convolutional} can be modelled as graphical data. In citation networks \cite{kipf2016semi}, papers can be viewed as nodes of a graph, and the link between different papers via citation can be modelled as edges of that graph. One of the basic differences of Euclidean and Non-Euclidean space is, unlike Euclidean space, a straight line joining any two points in Non-Euclidean space is not necessarily the shortest path between them. 
Therefore, applying the Artificial Neural Network technique to solve different task on the Non-Euclidean domain is not that straight forward and has several challenges \cite{wu2020comprehensive}.
\begin{itemize}
    \item Graph data are full of non-uniformity. Each graph consists of a different number of nodes, and each node in the graph has a variable number of neighbours. Hence, convolution type operations are not directly applicable here.
    \item Application of the existing Machine Learning algorithm is not always possible because instances are not independent here. In graphical data, the nodes share connections with other nodes via edges. So, to extract maximum information about the data while performing any specified task, it is necessary to capture this inter dependency among nodes.
\end{itemize}

The purpose behind Graph Neural Network is to perform various types of prediction and analytical tasks on graphs using deep learning approaches, as it is known that deep learning approaches are computationally capable of approximating most of the practically valuable functions on graphs \cite{scarselli2008computational}. The main idea behind 
Graph Neural networks is to find low-dimensional vector embedding of nodes in a large graph, which will capture the structural information of the graph as well as the feature information of the nodes.

However, previous works on Graph Neural Network are focused on the graphs, where the input feature space is countable \cite{xu2018powerful}. There are several instances where it is difficult to allocate a precise value to some features. Assigning an interval of value to one extend manages to capture the uncertainty of the feature space \cite{dutta2015some}. For example, rather than recording a person's weekly expense for every week during a specific period, it can be summarized to an interval with the minimum and maximum weekly outlay. An interval of systolic and diastolic pressure describes the blood pressure of a person  \cite{ahn2012resampling}. The traditional approach to analyze the interval-valued data was to use the midpoint of the interval to a regression model \cite{billard2000regression}.
Nevertheless, the Center Method (CM) does not consider the variations of the intervals. To overcome this drawback, Center-Range Method (CRM) uses two interval-valued regression models for mid-points and the ranges of the interval values \cite{neto2004univariate}. However, in general, the mid-points and half ranges are related, which is not taken into account by CRM \cite{ahn2012resampling}. Bivariate Center and Range Method (BCRM) uses two regression models using mid-points and half-ranges of the intervals, which take care of the effects of intervals widths \cite{billard2006symbolic}. \\
All these linear regression models can be used to analyze data with a linear pattern. Using Multi-layer perceptrons (MLPs)  \cite{hornik1989multilayer,hornik1991approximation}, a GNN model is proficient in overcoming this deficiency. Then an interesting question to be asked will be, why not use two end points of an interval as two separate features of a node and apply an existing GNN architecture accepting countable node feature of a graph. Answer to this question is although neural network has ability to capture relationship among different node features, interval is a quantitative measurement, where there is an order and the difference of two end points is meaningful. Our aim is to exploit this property of interval and develop an appropriate GNN architecture, where the model is capable enough to accept multiple intervals and output a single representation on its own. This will allow us to consider an interval as a unit through out the progress of the algorithm and perform the classification task as a function of the interval valued feature as well as the structure of the graph. Motivated by this situation, we introduce a new aggregation function of intervals, named as $agr_{new}$.
and develop a neural architecture Interval-valued Graph Neural Network (IV-GNN) to apply on interval-valued data using MLPs and the newly developed $agr_{new}$ as its basic building blocks. 

Now, we give a summary of our main contributions here:
\begin{enumerate}
    \item We introduce a new aggregation function $agr_{new}$ and precisely discuss its representational power.
    \item We identify interval structures that previously available interval aggregation schemes can not distinguish.
    \item We develop a neural-based architecture Interval-Valued Graph Neural Network (IV-GNN), that can deal with interval-valued features. 
    \item We discuss the space and time complexities of our proposed algorithm and validate our theoretical findings through performing graph classification tasks on several datasets.
\end{enumerate}

The rest of the paper is organized as follows. The related works on Graph Neural Network are reviewed in the Section \ref{related}. We have discussed about necessary basics of Graph Neural Network, Interval Mathematics and Aggregation Operator in the Section \ref{background}. The proposed framework for Interval-Valued Graph Neural Network (IV-GNN) is introduced in Section \ref{propframe}. The results of the experimental study are presented in Section \ref{experi}, and finally, the concluding remarks are made in Section \ref{conclu}.
\section{Related Works}\label{related}
Deep learning is a learning technique of the unknown structure of the input data distribution using multiple layers in the network \cite{bengio2012deep}. 
Encouraged by the commercial success of deep learning approaches in Euclidean structured data, a large number of methods has been introduced for Non-Euclidean data as well \cite{liu2020introduction}. Unlike Euclidean space, well-known properties like global parametrization, coordinate system, vector space structure of shift-invariance are not satisfied in Non-Euclidean space \cite{bronstein2017geometric}. Therefore, one of the key challenges was, to introduce the concept of convolution type operation in graphical data.
\newline
Graph Neural Network model approaches can be classified into two categories, namely spectral-based and spatial-based methods 
. 
The basic approach of spectral based GNN is to introduce filters from the graph signal processing perspective on a similar notion of the traditional convolutional neural network (CNN) \cite{shuman2013emerging}. 
However, due to high computational cost and lack of scalability, the relatively newer field of research, spatial-based GNN has gained popularity as they can handle large graphs by aggregating the neighbouring nodes. 
 We overview few models with spatial-based approach here,
\begin{enumerate}
    \item One of the earliest work in this area, Graph Neural Network (GNN) \cite{scarselli2008graph} recursively updates latent node representations by exchanging information with the neighbouring nodes, until equilibrium is reached. The recurrent function is chosen to be a contraction mapping to ensure convergence.
    \item Gated Graph Neural Networks (GGNN) \cite{cho2014learning} uses a gated recurrent unit as the recurrent function and use back-propagation through time (BPTT) for parameter learning. Hence, the condition on parameters to converge is no longer there, which reduces the number of steps.
    \item Stochastic Steady-State Embedding (SSE) \cite{dai2018learning} uses a recurrent function that takes a weighted average of the states from previous steps and a new state to ensure the stability of the algorithm.
    \item GraphSage \cite{hamilton2017inductive} proposes a batch-training algorithm, which improves scalability for large graphs. It samples a fixed-sized neighbourhood of a node to aggregate information.
    \item Graph Isomorphism Network GIN \cite{xu2018powerful} is one of the best performing models reported in the literature. It has been claimed that both the GIN and WL test are equally powerful in a graph classification task. It imposes a constraint on the functions used in the model to be injective to achieve the maximum representational power of a GNN.
    \item Higher order Graph Neural Network $k$-GNN \cite{morris2019weisfeiler} uses $k$-dimensional neighbourhood of a node to aggregate information from these and generate the low-dimensional representation for the mode. 
\end{enumerate}

All of these above discussed GNN-architectures have one limitation in common; that they can only accept countable features. The categorical features can also be handled by these models by converting them to integer data using integer encoding or one-hot encoding. However, there are several instances as mentioned earlier, where it is more convenient to represent the feature as interval and  treat interval as a single unit throughout the progress of the algorithm. Therefore, in this paper, a new interval aggregation scheme $agr_{new}$ has been introduced, satisfying all the necessary properties of a general aggregation function. Then, we have proposed Interval-Valued Graph Neural Network (IV-GNN) using $agr_{new}$ as aggregation operator, which can handle interval-valued features. We have performed graph classification tasks on four bioinformatics datasets, two social network datasets, and six synthetic datasets and have found that IV-GNN (with the proposed interval aggregation scheme) performs better than GNNs with various existing interval aggregation functions. Moreover, we demonstrate that our proposed model using degenerate interval-valued features (a special case of the more general IV-GNN model) outperforms other state-of-the-art approaches, that accept only countable features, for the graph classification task.
\section{Background and Definition}\label{background}
This section overviews the Graph Neural Network framework, basics of Interval Mathematics, and Aggregation Operators.
\subsection{Graph Neural Network.}

In this sub-section, we formally present the notion of Graph Neural Networks and the related concepts.
\newline
Let $G=(V,E)$ be a graph, where $x_{v}$ is the node feature vector associated a node $v\in V$. We want GNN to solve mainly two kinds of tasks.
\begin{itemize}
    \item \textbf{Node Classification } Let every node $v\in V$ has label $y_{v}$ associated with it. Then the objective is to get a vector representation $h_{v}$ for $v$ in Euclidean space such that $y_{v}$ turns out to be a function of $h_{v}$, i.e. $y_{v}=f(h_{v})$.
    \item \textbf{Graph Classification } Let $\{ G_{1},...,G_{N}\} $ be a collection of graphs, where $\{ y_{1},...,y_{N}\} $ is the set of associated labels of the graphs. Then our aim will be to learn a Euclidean representation $h_{G}$ for a particular graph $G$ such that $y_{G}$ turns out to be a function of $h_{G}$, i.e. $y_{G}=g(h_{G})$.
\end{itemize}
\begin{figure}[htbp]
\centerline{\includegraphics[width=0.5\textwidth]{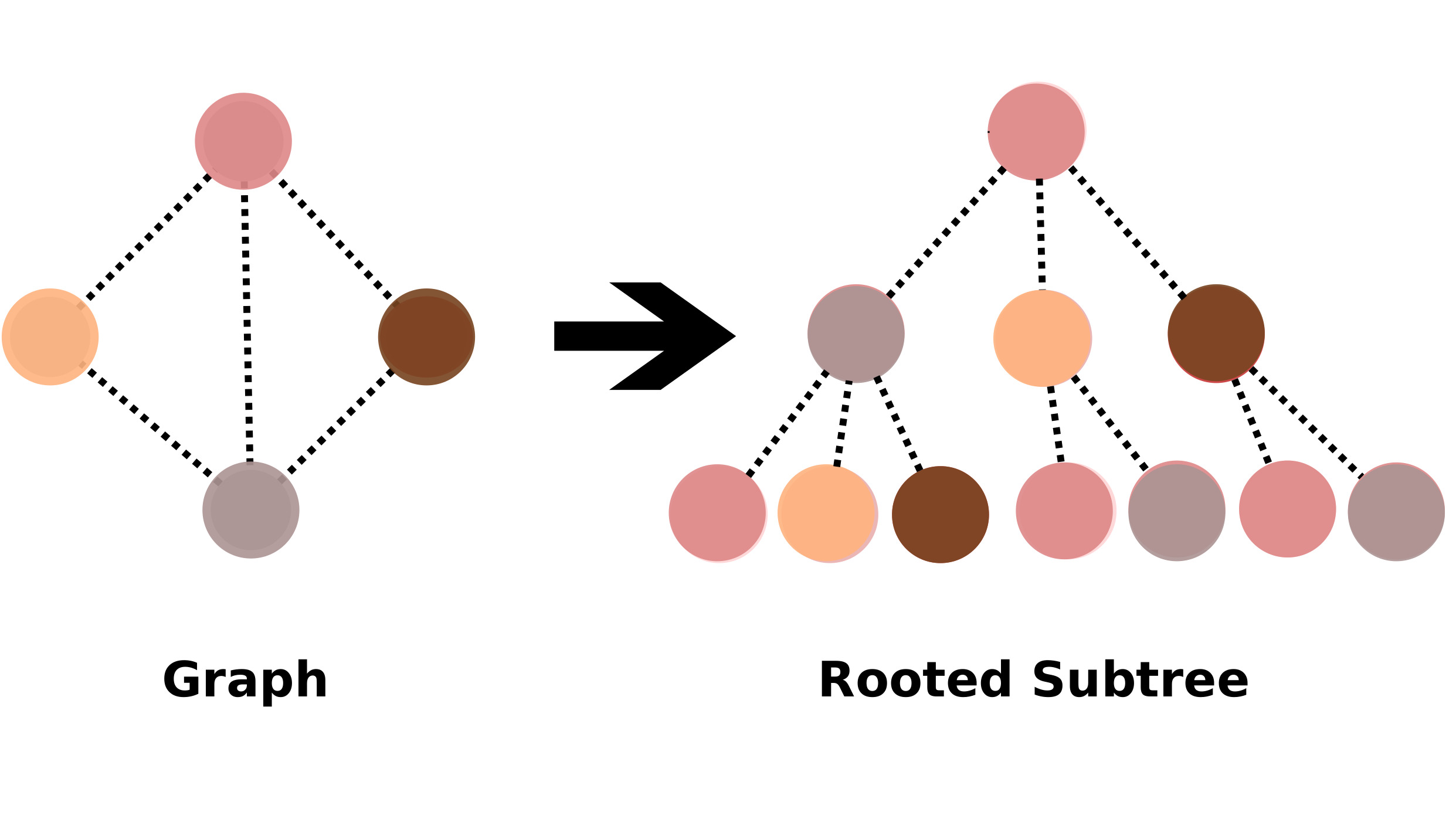}}
\caption{The basic framework behind GNN. The left picture shows the original graph. The right picture shows a 2-level rooted subtree structure at the pink node. After initializing at the leaf nodes, GNN can capture the feature information at the rooted node recursively. }
\label{fig}
\end{figure}
Now we describe the forward propagation algorithm (Algorithm \ref{Algorithm 1}) of any GNN model, which is used to generate embedding of nodes using graph structure and node features.

\begin{algorithm} [h!]
\caption{GNN framework for embedding generation}  
\label{Algorithm 1}
\KwIn{ Graph $ G(V,E) $ ; input features $ \{x_{v}, \forall v \in V \} $ ; depth K ; aggregator functions $AGGREGATE^{(k)}$ ,$\forall$k $\in \{1,2,...,K\}$ ; combine functions $COMBINE^{(k)}$ , $\forall$k $\in \{1,2,...,K\}$}
\KwOut{ Vector representations $z_{v}$ for all $v \in$ V}
$h_{v}^{0}\gets x_{v} ,  \forall v \in V$
\\
\For {$k=1,2,...,K$}{
    \For {$v\in V$}{
        $h_{N(v)}^{k}\gets AGGREGATE^{(k)}(\{h_{u}^{k-1};\forall u \in N(v)\})$
        $h_{v}^{k} \gets COMBINE^{(k)}(h_{v}^{k-1},h_{N(v)}^{k})$
    }
}
$z_{v} \gets h_{v}^{k}; \forall v \in V$
\\
\Return{$z_{v}$}
\end{algorithm}

In this algorithm, $N(v)$ stands for the neighbourhood of a vertex $v$. The main idea of GNN evolves around two functions $AGGREGATE^{(k)}$ and $COMBINE^{(k)}$. $AGGREGATE^{(k)}$ function is used to accumulate information from the neighbouring nodes, and $COMBINE^{(k)}$ function is used to update the existing representation vector of a node after $(k-1)$ iteration with the help of aggregated information from its neighbours. Here the iteration number stands for how many hop-neighbourhood of a node we want to consider to get its representation vector.
\newline
For graph classification, we have another function named READOUT, which will accept the representation of every node after the final iteration $K$ and predict the graph label.
\begin{equation*}
    h_{G}=READOUT(\{h_{v}^{K}|v\in V \})
\end{equation*}
Several AGGREGATE and COMBINE functions have been proposed in the GNN literature. In the GraphSAGE architecture \cite{hamilton2017inductive}, the AGGREGATE and COMBINE functions have been defined as follows,
\begin{equation}
    h_{v}^{k}=\sigma(W_{2}.CONCAT(h_{v}^{k-1},MAX (\{ReLU (W_{1}. h_{u}^{k-1}),\forall u \in N(v)\}))
\end{equation}
where $W_{1}$ and $W_{2}$ are learnable weight matrices, CONCAT represents vector concatenation. $\sigma$ is a non linear function.
\newline
In Graph Convolution Network \cite{kipf2016semi}, the AGGREGATE and COMBINE function can be defined as,
\begin{equation}
    h_{v}^{k}= ReLU (W. MEAN \{ h_{u}^{k-1},\forall u \in N(v)\cup \{v\}\})
\end{equation}

In Graph Isomorphism Network \cite{xu2018powerful}, the model uses a sum aggregator over max or mean aggregator due to its more discriminative power.
\newline
GIN updates existing node representations as,
\begin{equation}
    h_{v}^{k}=MLP^{k}((1+\epsilon^k).h_{v}^{k-1}+\Sigma_{u\in N(v)}h_{u}^{k-1})
\end{equation}
MLP represents a multi-layer perceptron, and $\epsilon$ can be a parameter that needs to be learned or assigned as a fixed scalar. Any spatial-based GNN can be at most as powerful as WL test \cite{xu2018powerful}. This network is one of those GNN models, which is equally powerful as the WL test distinguishing a broad class of graphs effectively and efficiently.
\newline

\subsection{Interval Mathematics}
The works discussed so far take care of the situation when the feature space is countable. Before generalizing this idea to interval-valued feature space, we introduce the interval mathematics and properties of any aggregating function. Note that, if the node feature is from countable space, it can be considered a degenerate interval, i.e. the start and end points of the interval are same.
\newline
Let us consider $\mathcal{U}=\{[a,b]|0\leq a \leq b \leq 1\}$ along with $\cap_{0}$ and $\cup_{0}$ defined by
\begin{equation*}
    [x_{1},x_{2}]\cap_{0}[y_{1},y_{2}] =[min(x_{1},y_{1}),min(x_{2},y_{2})]
\end{equation*}
    
\begin{equation*}
    [x_{1},x_{2}]\cup_{0}[y_{1},y_{2}]=[max(x_{1},y_{1}),max(x_{2},y_{2})]
\end{equation*}

($\mathcal{U},\cap_{0}, \cup_{0} $) forms a complete lattice.  A partially ordered set is said to be a complete lattice if all subsets have both a supremum (join) and an infimum (meet). We denote aggregation operator as $agr_{0}$ which is based on $\cap_{0}$.
\newline
Now to assign the value of the aggregation so that it takes care of every individual's opinions, the aggregated interval needs to be defined as an interval that lies in the intersection of everyone's opinion. If we define the aggregator function as $\cap_{0}$ as defined above, then the function is biased towards the interval with a lower value. Therefore, to overcome this drawback, in  \cite{dutta2015some},
($\mathcal{U},\cap_{e},\cup_{e}$) is defined as a lattice where the greatest lower bound, $\cap_{e}$ and the least upper bound, $\cup_{e}$ are defined as follows,
\begin{equation*}
\begin{split}
[x_{1},x_{2}]\cap_{e}[y_{1},y_{2}]&=[max(x_{1},y_{1}),min(x_{2},y_{2})],\\ &\text{ if }max(x_{1},y_{1})\leq min(x_{2},y_{2})\\  
&=[min(x_{2},y_{2}),min(x_{2},y_{2})] \text{  ,  otherwise }
\end{split}
\end{equation*}
\begin{equation*}
\begin{split}
[x_{1},x_{2}]\cup_{e}[y_{1},y_{2}]&=[max(x_{2},y_{2}),max(x_{2},y_{2})], \\
&\text{ if }x_{1}=x_{2},y_{1}=y_{2}  \\
&=[max(x_{1},y_{1}),max(x_{2},y_{2})], \\
&\text{ if }x_{1}=x_{2}<y_{1}<y_{2}  \\
&=[min(x_{1},y_{1}),max(x_{2},y_{2})] \text{  ,  otherwise } 
\end{split}
\end{equation*}
($\mathcal{U},\cap_{e},\cup_{e}$) is a complete lattice. We denote aggregation operator as $agr_{e}$ which is based on $\cap_{e}$.
\subsection{Aggregation Operators}
Any aggregation operator \cite{calvo2002aggregation} for fixed $n \geq 2$ is defined by a function $agr:[[0,1]\times[0,1]]^{n}\rightarrow[0,1]\times[0,1]$ fulfilling at least the two following axioms.
\begin{itemize}
    \item \textbf{Boundary conditions } 
    \begin{center}
     $agr(I_{min},I_{min},...,I_{min})=I_{min}$,
     
     $agr(I_{max},I_{max},...,I_{max})=I_{max}$,
    \end{center}where $I_{min}$ and $I_{max}$ are minimal and maximal possible inputs respectively.

    \item \textbf{Monotonic increasing }
    \newline
    $\forall (I_{1},I_{2},...,I_{n}),(J_{1},J_{2},...,J_{n}) \in [[0,1]\times[0,1]]^{n}$ such that $I_{i} \leq J_{i}, \forall i \in \mathbb{N}$ then 
    \begin{center}
        $agr(I_{1},I_{2},...,I_{n}) \leq agr(J_{1},J_{2},...,J_{n})$
    \end{center}
\end{itemize}
Beside these properties, we want our aggregations to satisfy two additional axioms.
\begin{itemize}
    \item \textbf{Symmetry }
    \newline
    $agr(I_{1},I_{2},...,I_{n})=agr(I_{p(1)},I_{p(2)},...,I_{p(n)})$ for any permutation $p$ on $\mathbb{N}^{n}$
    \item \textbf{Idempotency }
    \newline
    $agr(I,I,...,I)=I $,  $\forall I  \in [0,1]\times[0,1]$
\end{itemize}
\section{Theoretical Framework }\label{theoretical}
We introduce a new order relation $\subseteq_{new}$ on  $\mathcal{U}$ such that ($\mathcal{U},\subseteq_{new},\cap_{new},\cup_{new}$) forms a lattice and will give a comparable study against two previously discussed order relation.
\subsection{\textbf{Definition}}
$\subseteq_{new}$ is a binary relation on $\mathcal{U}$ defined as below
\begin{equation*}
    [x_{1},x_{2}]\subseteq_{new}[y_{1},y_{2}] \text{ if } (y_{1} < x_{1}) \text{ or } (x_{1}=y_{1} \text{ and }x_{2} \leq y_{2}).
    \end{equation*}

\subsection{\textbf{Proposition}   \texorpdfstring{($\mathcal{U},\subseteq_{new}$)} \text{ forms a poset}.}
To show ($\mathcal{U},\subseteq_{new}$) forms a poset, we have to show the relation $\subseteq_{new}$ is reflexive, antisymmetric, and transitive.
\begin{itemize}
    \item{Reflexivity.}
    \begin{equation*}
     [x_{1},x_{2}]\subseteq_{new} [x_{1},x_{2}] \text{ as } x_{1} = x_{1} \text{ and } x_{2} \leq x_{2}   
    \end{equation*}
    \item{Antisymmetricity.}
    \begin{equation*}
    \begin{split}
    [x_{1},x_{2}]\subseteq_{new}[y_{1},y_{2}] \implies& y_{1} < x_{1} \text{ or }\\
    &x_{1}=y_{1} \text{ and }x_{2} \leq y_{2} ....(i)
    \end{split}
    \end{equation*}
    \newline
    \begin{equation*}
    \begin{split}
     [y_{1},y_{2}]\subseteq_{new}[x_{1},x_{2}] \implies& x_{1} < y_{1} \text{ or }\\ 
     &y_{1}=x_{1} \text{ and }y_{2} \leq x_{2} ....(ii) 
    \end{split}
    \end{equation*}
    \newline
    (i) and (ii) imply 
    \begin{equation*}x_{1}=y_{1} \text{ and } x_{2}=y_{2}.
    \end{equation*}
    Hence , 
    \begin{equation*}
        [x_{1},x_{2}]= [y_{1},y_{2}]
    \end{equation*}
    
    \item Transitivity can be shown similarly.
    
\end{itemize}

\subsection{\textbf{Proposition}   (\texorpdfstring{$\mathcal{U},\subseteq_{new}$ )}\text{ forms a lattice.}}
($\mathcal{U},\subseteq_{new}$) is a lattice where the greatest lower bound, say $\cap_{new}$ and the least upper bound, say $\cup_{new}$ are defined as follows,
\begin{equation*}
\begin{split}
[x_{1},x_{2}]\cap_{new}[y_{1},y_{2}]&=[max(x_{1},y_{1}),min(x_{2},y_{2})] \\
&\text{ if }max(x_{1},y_{1}) \leq min(x_{2},y_{2}) \\
&\leq max(x_{2},y_{2}) \neq 1 \\   
&=[max(x_{1},y_{1}),1] \text{  ,  otherwise }\\
    [x_{1},x_{2}]\cup_{new}[y_{1},y_{2}]&=[min(x_{1},y_{1}),max(x_{2},y_{2})] \\
    &\text{ if }max(x_{2},y_{2}) \neq 1    \\
    &=[min(x_{1},y_{1}),min(x_{2},y_{2})] \text{  ,  otherwise }
\end{split}
\end{equation*}

\subsection{\textbf{Proposition}   (\texorpdfstring{$\mathcal{U},\subseteq_{new}$)}\text{ forms a bounded lattice.}}
For an arbitrary interval $[x,y]\in \mathcal{U}$, we have $0 \leq x \leq y \leq 1$. Hence $[1,1] \subseteq_{new} [x,y] \subseteq_{new} [0,1] $.
\newline
We denote aggregation operator as $agr_{new}$, which is based on $\cap_{new}$ .
\subsection{\textbf{Proposition}   \texorpdfstring{$agr_{new}$} \text{ satisfies all four conditions of aggregation function.}}
$agr_{new}$ satisfies boundary conditions where $[1,1]$ and $[0,1]$ are the minimal and maximal possible inputs respectively.
\section{Interval-Valued Graph Neural Network (IV-GNN)}\label{propframe}
We develop a general GNN model where the feature space need not be countable. Releasing this constraint on the feature space, our model can capture the graph's structural properties and can extract useful information from interval-valued features.
Therefore, our proposed architecture is much more general in nature and to the best of our knowledge, no existing models of GNN in literature can accept nodes' feature, which are intervals.
\subsection{AGGREGATE and UPDATE function of IV-GNN}
As we have already discussed that spatial-based GNN architecture has two primary functions, namely $AGGREGATE$ and $COMBINE$, we use our newly develop interval aggregation operator \texorpdfstring{$agr_{new}$ }        \text{function} to aggregate the neighbouring nodes' embedding to extract maximum information out of it.
We develop our model with aggregation and update function on $k$-th iteration defined as
\begin{equation}
    h_{v}^{k}=\Phi(h_{v}^{k-1} , agr_{new}(\{h_{u}^{k-1} : u\in N(v)\}))
\end{equation}, where $\Phi$ is the update function. To choose $\Phi$, we take the help of the Universal Approximation Theorem  \cite{hornik1989multilayer,hornik1991approximation}, which states that, using the multi-layer feed-forward architecture in a neural network framework, makes it a universal approximator of any continuous functions under mild assumptions on the activation function \cite{liu2011hard}. However, the limitation of continuity of the function was released much later. It has been shown that, a single hidden layer feed-forward neural network (SLFNNs) can approximate any real, piece wise continuous function almost uniformly  \cite{llanas2008constructive}.
\newline
Therefore, IV-GNN has the updating step as,
\begin{equation} \label{equation2}
    h_{v}^{k}=MLP^{k}(agr_{new}((1+\epsilon^{k})h_{v}^{k-1},agr_{new}\{h_{u}^{k-1}:u\in N(v)\}))
\end{equation}
\subsection{Graph-Level READOUT function of IV-GNN}
The purpose of this graph-level read-out function is to get embedding of the graph using the embedding of the nodes. If we want to perform jobs like node classification or link prediction within a graph then the node embedding using aggregation and update function at node level are sufficient.
\newline
While selecting the Graph-level READOUT function, we want to focus on the following aspects.
\begin{itemize}
    \item We want to use the structural information/node embedding that we have got after every iteration. It may so happen that the node embedding from an earlier iteration captures more information about the graph rather than the final iteration.
    \item The graph-level function should be injective to have our GNN variant as powerful as WL-test of isomorphism by distinguishing between different structures/node features. 
\end{itemize}
 Hence, to achieve a skip connection like architecture similar to Jumping Knowledge \cite{xu2018representation} and maximal distinguishing power, we concatenate the SUM of the node embeddings after every iteration.
    \begin{equation}\label{equation3}
        h_{G}=CONCAT(SUM (\{h_{v}^{k}|v\in G\})|
        k=0,1,...,K)
    \end{equation}

\subsection{Challenging structures for \texorpdfstring{$agr_{0}$ and $agr_{e}$} .}
\begin{figure}[htbp]
\centerline{\includegraphics[width=0.5\textwidth]{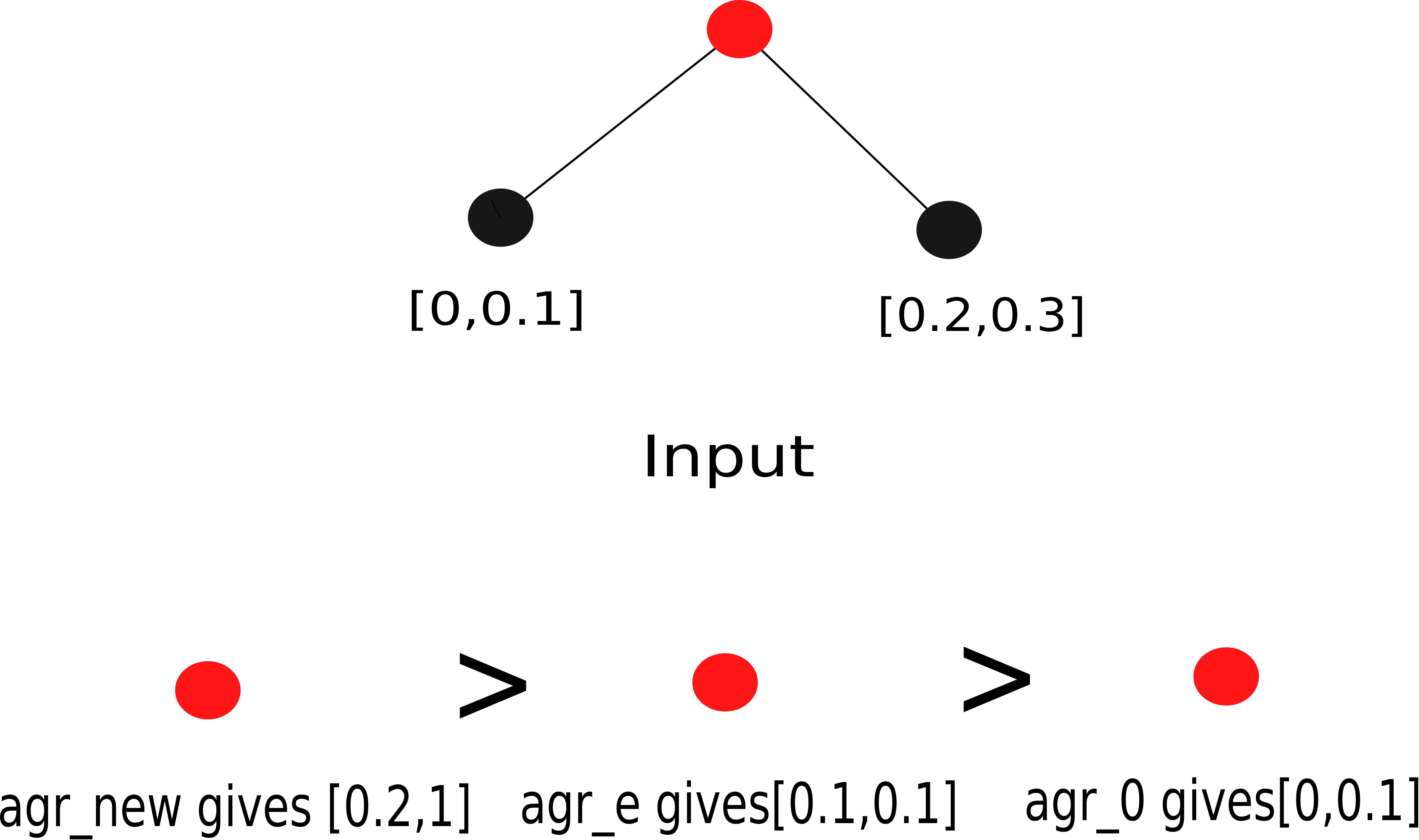}}
\caption{Ranking by expressive power for $agr_{0}$, $agr_{e}$, $agr_{new}$. Among these three aggregators, $agr_{new}$ has the maximum ability to capture the structural and feature related information. One thing to notice that, the right end point of an resulting interval equals to 1 expresses the fact that two aggregating intervals are non overlapping. $agr_{e}$ is equally powerful as $agr_{new}$ when two intervals have non-null intersection. $agr_{0}$ captures the smaller interval (according to the order relation) and ignores the other interval.}
\label{rankingofinterval}
\end{figure}

\begin{figure}[htbp]
\centerline{\includegraphics[width=0.4\textwidth]{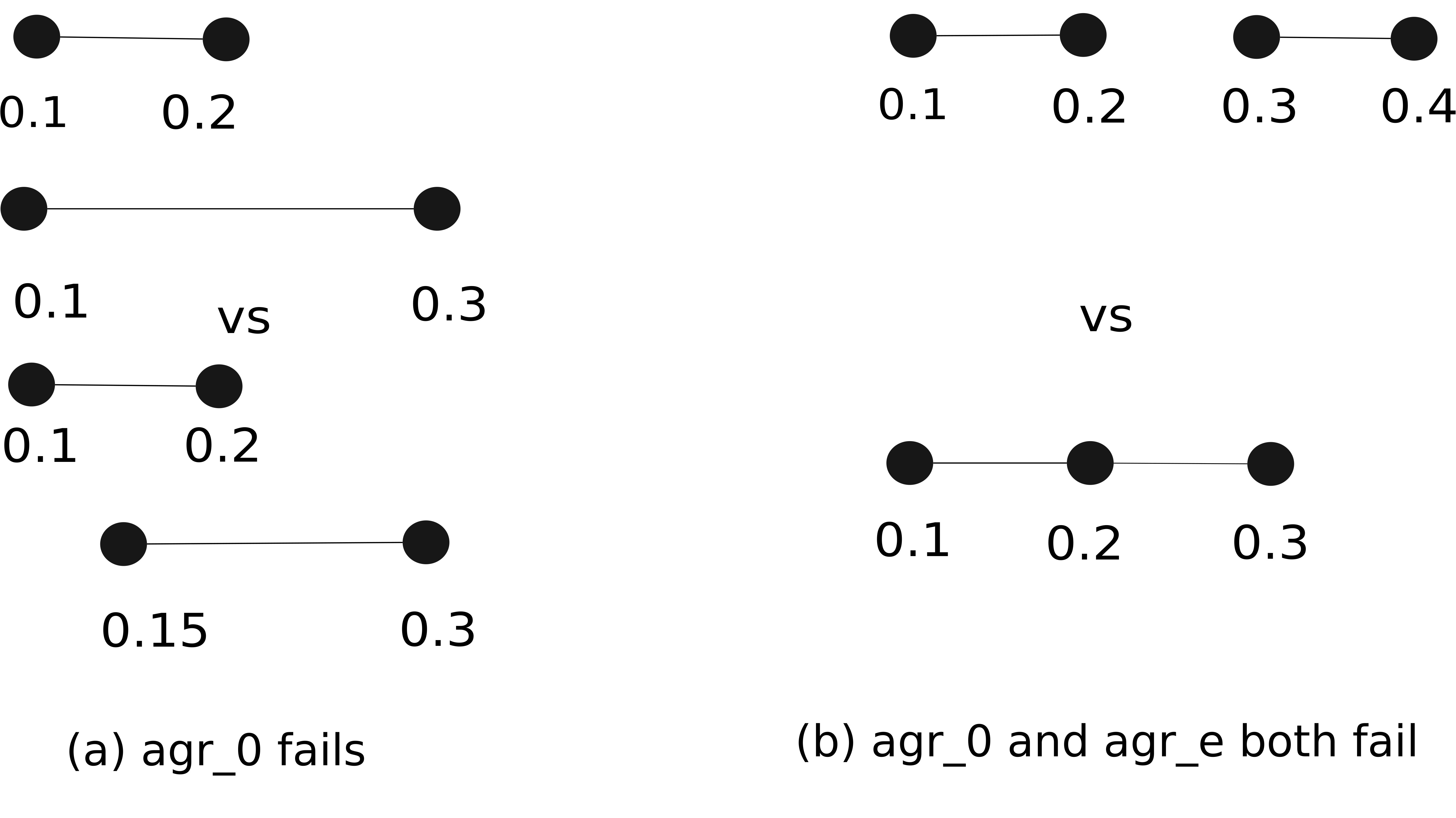}}
\caption{\textbf{Examples of interval structures that $agr_{0}$ and $agr_{e}$ fail to distinguish.} In the left picture, $agr_{0}$ is giving same aggregated interval even though two sets of intervals are different. In the right picture, $agr_{0}$ and $agr_{e}$, both are unable to recognise the differences between the two sets of intervals.}
\label{exampleofinterval}
\end{figure}

The main idea of choosing a powerful aggregator is to capture and compress the amount of structural and feature information from the nodes in its aggregated output value. Also, the aggregator function should be permutation invariant. That is, the order of the interval during aggregation should be immaterial. In this context, $agr_{0}$ , $agr_{e}$ and $agr_{new}$, all are satisfying this condition. In Fig. \ref{rankingofinterval}, we have shown the ranking of three aggregation functions pictorially with respect to their representational power. We have denoted the root node as the red node and the adjacent nodes of the root node as the black node whose features need to be aggregated and combined with the root node. In Table. \ref{intervaltabel}, we have illustrated these facts with the help of examples.
\newline
In the Fig. \ref{exampleofinterval}a, we have three intervals, $I_{1}=[0.1,0.2]$ , $I_{2}=[0.1,0.3]$ and $I_{3}=[0.15,0.3]$. We construct two sets of intervals $S_{1}$ and $S_{2}$, where $S_{1}=\{I_{1},I_{2}\}$ and  $S_{2}=\{I_{1},I_{3}\}$, which need to be aggregated. Now,
\begin{equation*}
    agr_{0}(S_{1}) = agr_{0}(S_{2}) = I_{1}
\end{equation*}
Hence, in this case $agr_{0}$ fails to capture the desired information about the intervals. However, $agr_{e}$ and $agr_{new}$ will give the aggregated intervals as the intersecting sub intervals.
\begin{equation*}
    agr_{e}(S_{1}) = agr_{new}(S_{1}) = [0.1,0.2]
\end{equation*}
\begin{equation*}
    agr_{e}(S_{2}) = agr_{new}(S_{2}) = [0.15,0.2]
\end{equation*}
\newline
In Fig. \ref{exampleofinterval}b, two sets of intervals need to be aggregated $S_{3}=\{I_{1},I_{4}\}$ and $S_{4}=\{I_{1},I_{5}\}$, where $I_{4}=[0.3,0.4]\}$ and $I_{5}=[0.2,0.3]$. Here also, $agr_{0}$ will fail to distinguish between them.
\begin{equation*}
    agr_{0}(S_{1}) = agr_{0}(S_{2}) = I_{1}
\end{equation*}
But according to definition, $agr_{e}$ will give same degenerate interval $[0.2,0.2]$ for both $S_{3}$ and $S_{4}$. However $agr_{new}$ can differentiate between them as it will aggregate and give the resultant intervals as $[0.3,1]$ and $[0.2,0.2]$ for $S_{3}$ and $S_{4}$ respectively.
\begin{equation*}
    agr_{e}(S_{3}) = agr_{e}(S_{4}) = [0.2,0.2]
\end{equation*}
\begin{equation*}
\begin{split}
    agr_{new}(S_{3}) &= [0.3,1] ,\\
    agr_{new}(S_{4}) &= [0.2,0.2]
\end{split}
\end{equation*}
As $I_{1}$ and $I_{4}$ are non-intersecting, $agr_{new}$ will be able to capture the uncertainty and express it by assigning a broader interval.
\begin{table}[htbp] 
\caption{Comparison between three interval aggregator.}
\begin{center}
\begin{tabular}{|c|c|c|c|c|}
\hline
Interval 1 & Interval 2 & \textbf{$agr_{0}$}& \textbf{$agr_{e}$}& \textbf{$agr_{new}$} \\
\hline
[0.1,0.2] & [0.1,0.3] & [0.1,0.2] & [0.1,0.2] & [0.1,0.2]\\
\hline
[0.1,0.2] & [0.15,0.3] & [0.1,0.2] & [0.15,0.2] & [0.15,0.2]\\
\hline
[0.1,0.2] & [0.3,0.4] & [0.1,0.2] & [0.2,0.2] & [0.2,1]\\
\hline
[0.1,0.2] & [0.2,0.3] & [0.1,0.2] & [0.2,0.2] & [0.2,0.2]\\
\hline
\end{tabular}
\end{center}
\label{intervaltabel}
\end{table}
\newline
The $agr_{e}$ will perform well if the two intervals have a non-null intersection. So, whenever the node features are not very diverse, $agr_{e}$ will be as powerful as the $agr_{new}$ aggregator.
\subsection{Model Training}
To estimate model parameters of IV-GNN, we need to specify an objective function to optimize. Since the task we focus on in this
work is Graph Classification task, loss is computed as the sum of the difference between the actual graph class and the predicted graph class for the graphs in the dataset.
\subsection{Memory and Space Complexity of training the embedding generation process of IV-GNN}
In order to find the worst case scenario, we assume that all $||E||$ edges are connected to all $||V||$ nodes of the graph $G$. As IV-GNN is a full-batch gradient descent process, it requires to store all the embeddings found from the intermediate layers, which requires $O(Kn)$ storage for one node. Here, $K$ denotes the number of layers and $n$ denotes the dimension of the embedding space. For the sake of simplicity, we keep the embedding space dimension same for every layer. Furthermore, at every layer, a weight matrix of size $n\times n$ is involved, which includes $O(Kn^{2})$ storage in total. Therefore, overall IV-GNN has a space-complexity of $O(K||V||n+Kn^{2})$.
\\
Now, we illustrate the time-complexity of our proposed model. As discussed previously, IV-GNN stores intermediate embeddings of every node, generated from each lower layer. In contrary to mini-batch algorithm like GraphSAGE \cite{hamilton2017inductive}, IV-GNN utilizes those saved embeddings and reuses those in the upper layer. Therefore, at every layer, previously layers' embeddings are multiplied with the weight matrix of size $n\times n$, which include $n^{2}$-many multilications, followed by some element-wise operations. Therefore, as a whole, for $K$ many layers and $||V||$ many nodes, IV-GNN has time complexity $O(K||V||n^{2}+K||E||n)$.   
\section{Experiments}\label{experi}
This section discusses the dataset used for experiments, and we evaluate our theoretical findings by comparing the training and test set performances of IV-GNN on the synthetic and real-life datasets.
\subsection{Datasets}
We have used six synthetic datasets, four bioinformatics datasets and two social network datasets to demonstrate the efficiency of our model. 
\begin{enumerate}
    \item Synthetic Datasets:  Our basic idea is to generate random graphs with a various number of nodes. Then based on two topological properties, we give tag and feature interval to the nodes and classify every graph in the datasets. The summary statistics of these synthetic datasets are provided in Table-\ref{tabelsynthetic}. The topological properties are listed below, 
    \begin{itemize}
        \item \textbf{Density } The density of a graph is defined as the ratio of the number of edges and the number of nodes \cite{lawler2001combinatorial}, i.e.,
        \begin{equation*}
            \text{density} (G) = \frac{|E|}{|V|}
        \end{equation*}
        where $G=(V,E)$ denotes a graph, $V$ is the set of nodes and $E$ is the edge set of the graph $G$.
        Average density of the dataset $D=\{G_{i},i=1,...,n\}$ can be calculated as, 
        \vspace{0.25cm}
        \begin{center}
            $avg\_density=\frac{\sum_{i=1}^{n}{density(G_{i})}}{n}$
        \end{center}
        We assign,
        \begin{equation*}
        \text{Graph class (G)}=\begin{cases}
        1, & \text{if density }(G) < \text{avg\_density}.\\
        0, & \text{otherwise}.
         \end{cases}
      \end{equation*}
      We assign, 
      \begin{equation*}
        \text{tag}(v)=\begin{cases}
        1, & \text{if degree }(v) < \text{average degree}.\\
        0, & \text{otherwise}.
         \end{cases}
      \end{equation*}
      where average degree is calculated over all nodes in the dataset.
      \newline
      To assign the interval valued feature to a node $v$, we follow the following rule,
      \begin{equation*}
        \text{feature interval}(v)=\begin{cases}
        [d_{min},d_{max}], & \text{if} |N(v)| >1\\
        [-1,d_{max}], & \text{if} |N(v)| =1\\
        [-1,0], & \text{otherwise} .
         \end{cases}
      \end{equation*}
      where
      \begin{center}
          $d_{min}$=\(\displaystyle\min_{u\in N(v)}degree(u)\),
      \end{center}
      \begin{center}
          $d_{max}$=\(\displaystyle\max_{u\in N(v)}degree(u)\)
      \end{center}
      \begin{figure}[htbp]
\centerline{\includegraphics[width=0.3\textwidth]{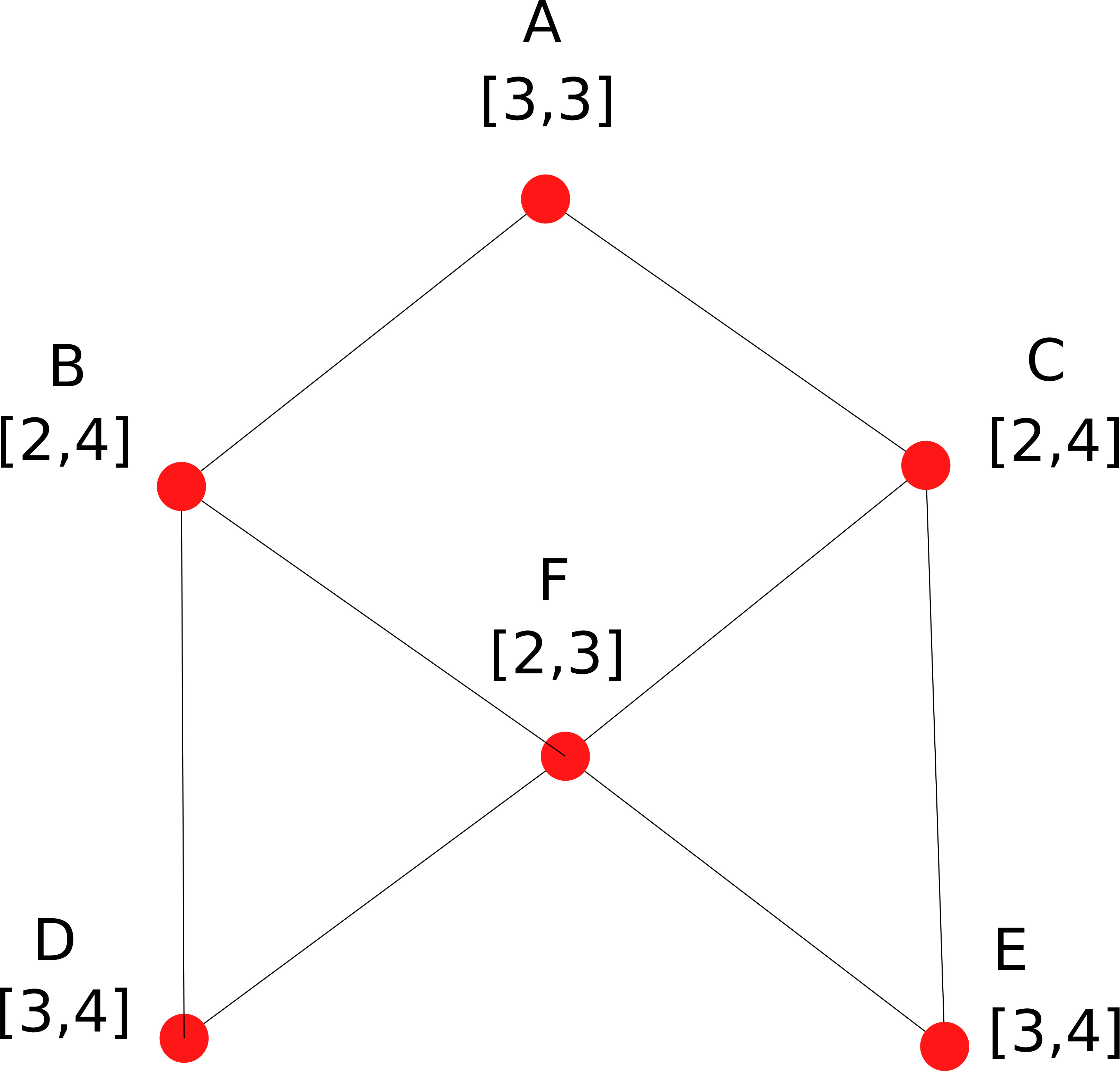}}
\caption{Explanation of assigning feature interval to a node. For the node A, it has two neighbours of degree. Hence the feature interval assigned to A is [3,3]. Similarly, node F has 4 neighbours of degree 2, 2, 3, 3. Hence the assigned to F is [2,3].}
\label{graph_example}
\end{figure}

      We have created three datasets, SYNTHETIC\_1\_200, SYNTHETIC\_1\_1000, SYNTHETIC\_1\_2000, with 200, 1000, 2000 graphs, respectively, according to this construction.
      
      \item \textbf{Average clustering coefficient  }The clustering coefficient $c(u)$ of a node $u$,is a measure of the likelihood that any two neighbors of $u$ are connected \cite{li2011graph}. Mathematically, the clustering coefficient of a node $u$ can be formulated as:
      \begin{equation*}
       c(u) = \frac{\lambda(u)}{\tau(u)}.   
      \end{equation*} where $\lambda(u)$ is the number of triangles engaging the node $u$. By triangle, we mean complete graph with three nodes and \begin{equation*}
          \tau(u)=\frac{degree(u)(degree(u)-1)}{2},
      \end{equation*} i.e.,  the number of triples a node $u$ has.
      \newline 
      In other words, the clustering coefficient for node $u$ is the ratio of the actual number of edges between two nodes from the neighbours of $u$ and the maximally possible numbers of edges between them. The clustering coefficient $C(G)$ of a graph $G$ is the average of $c(u)$ taken over all the nodes in the graph, i.e., 
      \begin{equation*}
          C(G) = \frac{1}{n}\sum_{i=1}^{n} c(u_{i})
      \end{equation*} 
      Average clustering coefficient of the dataset $D=\{G_{i},i=1,...,n\}$ can be calculated as, 
        \vspace{0.25cm}
        \begin{center}
            $avg\_cluster=\frac{\sum_{i=1}^{n}{C(G_{i})}}{n}$
        \end{center}
        We assign,
        \begin{equation*}
        \text{Graph class}(G)=\begin{cases}
        1, & \text{if } C(G) < avg\_cluster\\
        0, & \text{otherwise}.
         \end{cases}
      \end{equation*}
      We use node's degree as its tag.
      \begin{equation*}
        \text{tag}(v)= degree(v)
      \end{equation*}
      To assign the interval valued feature to a node $v$, we follow the following rule,
      \begin{equation*}
        \text{feature interval}(v)=[c_{min},c_{max}]
      \end{equation*}
      where
      \begin{center}
          $c_{min}$=\(\displaystyle\min_{u\in N(v)}c(u)\),
      \end{center}
      \begin{center}
          $c_{max}$=\(\displaystyle\max_{u\in N(v)}c(u)\)
      \end{center}
      We have created three datasets SYNTHETIC\_2\_200, SYNTHETIC\_2\_1000, SYNTHETIC\_2\_2000 with 200, 1000, 2000 graphs respectively according to this construction.
       
    \begin{table}[htbp] 
\caption{Statistics of the synthetic datasets }
\begin{center}
\begin{tabular}{|c|c|c|c|c|}
\hline
\textbf{Dataset} & \textbf{\textit{Size}}& \textbf{\textit{Classes}}& \textbf{\textit{Avg. nodes}}& \textbf{\textit{labels}} \\
\hline
SYNTHETIC\_1\_200 & 200 & 2 & 19.94 & 2\\
\hline
SYNTHETIC\_1\_1000 & 1000 & 2 & 19.83 & 2\\
\hline
SYNTHETIC\_1\_2000 & 2000 & 2 & 19.92 & 2\\
\hline
\hline
SYNTHETIC\_2\_200 & 200 & 2 & 19.94 & 23\\
\hline
SYNTHETIC\_2\_1000 & 1000 & 2 & 19.83 & 25\\
\hline
SYNTHETIC\_2\_2000 & 2000 & 2 & 19.92 & 25\\
\hline
\end{tabular}
\end{center}
\label{tabelsynthetic}
\end{table}
 \end{itemize}
    \item Bio-informatics datasets: 4 datasets MUTAG, PROTEINS, PTC and NCI1 \cite{yanardag2015deep} have been used for our experiment. The summary statistics of these bioinformatic datasets are provided in Table-\ref{tabelbioinformatic}.
    \begin{itemize}
        \item MUTAG consists of 188 graphs which have 7 discrete node labels. Each graph in the dataset represents a chemical compound \cite{debnath1991structure}.
        \item In the dataset PROTEINS, nodes represent secondary structure elements (SSEs) and two nodes share an edge if they appear as adjacent in the amino-acid sequence. Graph nodes have 3 different labels such as helix, sheet or turn \cite{borgwardt2005protein}.
        \item PTC includes 344 chemical compounds that describes the carciogenicity for male and female rats having 19 discrete node labels \cite{toivonen2003statistical}.
        \item NCI1 is a balanced dataset with 4100 nodes with 37 discrete labels, published by the National Cancer Institute (NCI). It contains chemical compounds, that are found to have the ability to suppress or inhibit the growth of a panel of human tumor cell lines \cite{wale2008comparison}.
        \begin{table}[htbp] 
\caption{Statistics of the Bioinformatics datasets used}
\begin{center}
\begin{tabular}{|c|c|c|c|c|}
\hline
\textbf{Dataset} & \textbf{\textit{Size}}& \textbf{\textit{Classes}}& \textbf{\textit{Avg. nodes}}& \textbf{\textit{labels}} \\
\hline
MUTAG & 188 & 2 & 17.9 & 7\\
\hline
PTC & 344 & 2 & 25.5 & 19\\
\hline
PROTEINS & 1113 & 2 & 39.1 & 3\\
\hline
NCI1 & 4110 & 2 & 29.8 & 37\\
\hline
\end{tabular}
\end{center}
\label{tabelbioinformatic}
\end{table}
    \end{itemize}
    \item Social network datasets: 2 datasets, IMDB-BINARY and COLLAB \cite{yanardag2015deep} have been used for our experiment.
    The summary statistics of these datasets are provided in Table-\ref{tabelsocialnetwork}.
    \begin{itemize}
        \item Movie Collaboration Dataset: IMDB-BINARY is a dataset of movie collaboration, wherein in each graph, nodes represent actors/actresses and two nodes share an edge if two actors/actresses act in the same movie. There are two graph classes Action and Romance genres.
        \item Scientific collaboration dataset: COLLAB is a dataset of scientific collaboration, acquired from 3 public collaboration datasets, namely High Energy Physics, Condensed Matter Physics and Astro physics \cite{leskovec2005graphs}. In  \cite{shrivastava2014new}, ego-networks of various researchers from each field has been generated, and each graph has a label according to the field of the researcher. Now the task will be to determine an ego-collaboration graph's label of a researcher. 
    \end{itemize}
\end{enumerate}
\begin{table}[htbp] 
\caption{Statistics of the Social network datasets used}
\begin{center}
\begin{tabular}{|c|c|c|c|c|}
\hline
\textbf{Dataset} & \textbf{\textit{Size}}& \textbf{\textit{Classes}}& \textbf{\textit{Avg. nodes}}& \textbf{\textit{labels}} \\
\hline
IMDBBINARY & 1000 & 2 & 19.8 & - \\
\hline
COLLAB & 5000 & 3 & 74.5 & - \\
\hline
\end{tabular}
\end{center}
\label{tabelsocialnetwork}
\end{table}
\subsection{Performance of IV-GNN}
Our goal is to allow the model to capture structural information and feature information from the graph. Therefore, we like to evaluate our model IV-GNN on Graph Classification task with interval-valued features of the nodes.
\subsubsection{Data Preparation}
To convert the feature values into intervals, we prepare the data as follows: rather than using the tag of a node as a node feature, we give a bias of $k_{1}$ and $k_{2}$, where $k_{1}$ and $k_{2}$ are selected from the normal distribution $N(0,1)$ to generate an interval-valued feature for that particular node. For example, if in a graph, a node has a tag as c, then we assign $[c-k_{1},c+k_{2}]$ as its feature interval.
\subsubsection{Baselines}
We evaluate our IV-GNN by comparing it with other frameworks with different interval aggregation operators. As there is no existing model, who can handle interval-valued features of the nodes, we use existing interval aggregation operators with same neural architecture as of IV-GNN, which will show expressive power of $agr_{new}$ against that of others experimentally. The details of the baselines are discussed below.
\newline 
\begin{itemize}
    \item \textbf{$agr_{0}$-based GNN } In this model, we choose $agr_{0}$ for interval-aggregation and $SUM$ as Graph-Level $READOUT$ function.
    \item \textbf{$agr_{e}$-based GNN } In this model, interval-aggregation operator $agr_{e}$ has been used as $AGGREGATE$ function. Like before, we use $SUM$ as the Graph-Level $READOUT$ function.
\end{itemize}
 We report the training set performance and test set performance in the figure \ref{training} and table \ref{table1} respectively.
\subsection{Performance with degenerate interval: A special case of IV-GNN}
In order to examine the performance of IV-GNN for countable features, we compare IV-GNN with respect to the state-of-the-art approaches which accept countable feature-value. For this experiment, we treat the exact value of the feature as a degenerate interval, i.e., we use exact feature value as same starting and end points of the interval. We report the training set performance and test set performance in the figure \ref{figuredegenerate} and table \ref{table2} respectively.
\subsubsection{Baselines}
We compare IV-GNN with \emph{three}, state-of-the-art GNN models, as briefly discussed below.
\begin{itemize}
    \item GraphSage \cite{hamilton2017inductive}: This uses a sampled neighbourhood and aggregates information from these to generate the Euclidean representation of the nodes.
    \item GCN \cite{kipf2016semi}: This model uses the MEAN-pool of the information from the neighbouring nodes and then combines the changes with the existing feature of a node using a nonlinear function.
    \item GIN \cite{xu2018powerful}: This model uses SUM as its aggregation function because of its maximal discriminative power and a nonlinear function as a combine function.
\end{itemize}
\subsection{Parameter Settings}
For IV-GNN, we adopt the parameter settings based on our hyperparameter study [see \ref{hyper}], and details are as follows,
\begin{itemize}
\item We have used $5$-layers of each GNN block where every MLP will have $2$-layers excluding the input layer. 
\item Each hidden layer has $\{32,128\}$ hidden units. We have used Batch Normalization in each hidden layer.
    \item We have used Adam optimizer \cite{kingma2014adam} with the initial learning rate 0.01 and decay the learning rate by 0.5 after every 50 epochs.
    \item Input batch size of training is $\{16,64\}$.
    \item The final layer dropout is 0.5 \cite{srivastava2014dropout}.
\end{itemize}
As recommended for GIN, we perform 10-fold cross-validation with LIB-SVM \cite{yanardag2015deep,chang2011libsvm,niepert2016learning} and perform the experiment for $350$ epochs.
\subsection{Results} 
\subsubsection{Training set performance }
\begin{figure}
  \begin{minipage}{.3\textwidth}
  \centering
  \includegraphics[width=\linewidth]{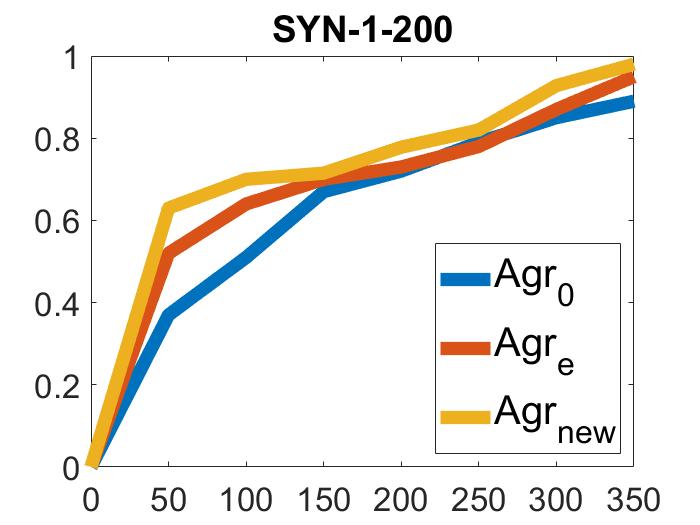}
  \end{minipage}
  \hfill
  \begin{minipage}{.3\textwidth}
  \centering
  \includegraphics[width=\linewidth]{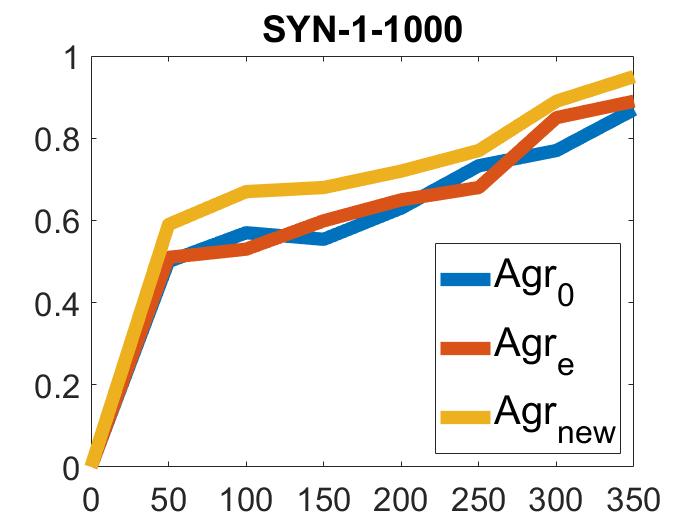}  
  \end{minipage}
  \hfill
  \begin{minipage}{.3\textwidth}
  \centering
 \includegraphics[width=\linewidth]{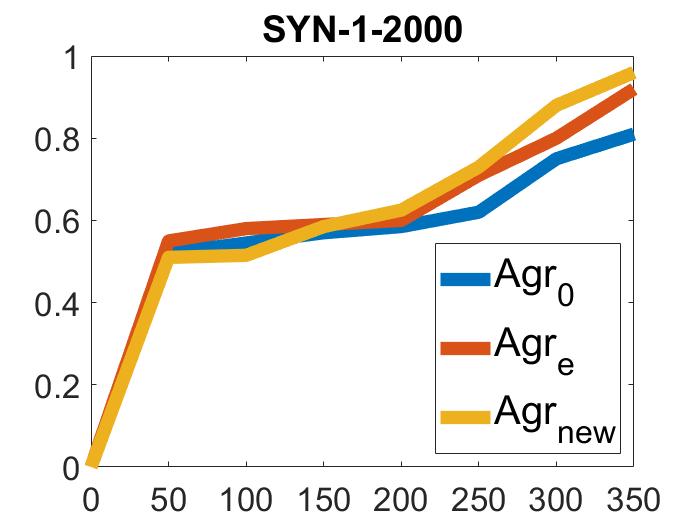}  
  \end{minipage}
  \newline
   \begin{minipage}{.3\textwidth}
   \centering
   \includegraphics[width=\linewidth]{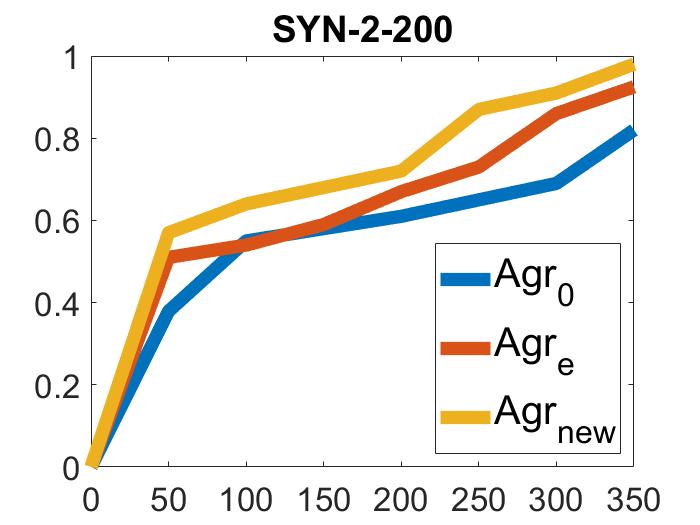}
  \end{minipage}
  \hfill
  \begin{minipage}{.3\textwidth}
  \centering
  \includegraphics[width=\linewidth]{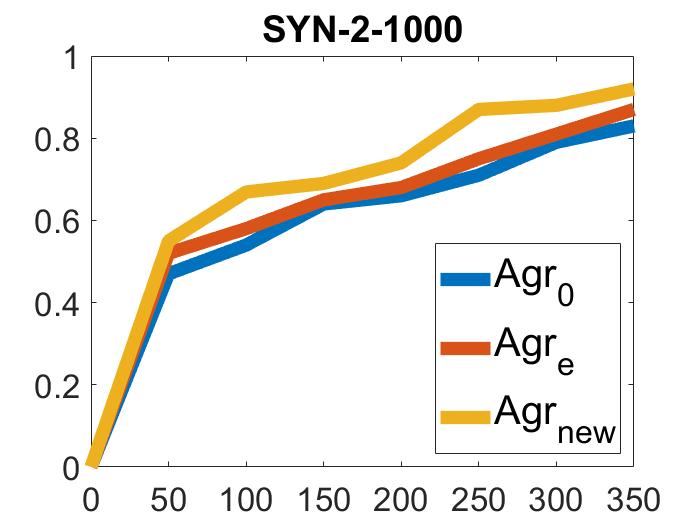}
  \end{minipage}
  \hfill
   \begin{minipage}{.3\textwidth}
  \centering
  \includegraphics[width=\linewidth]{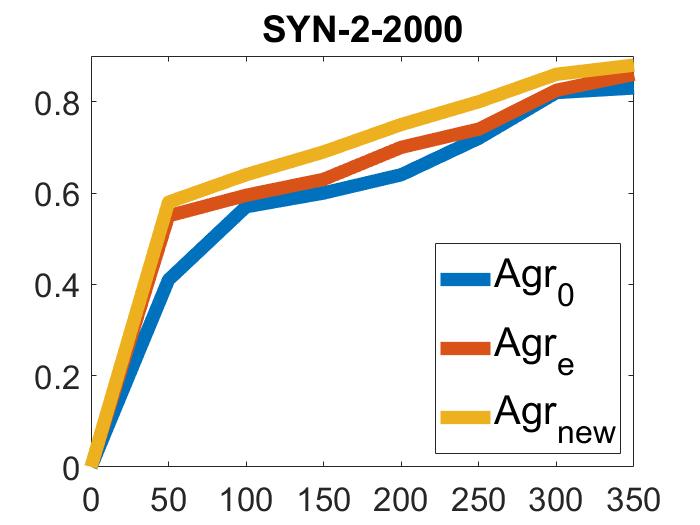}
  \end{minipage}
  \newline
  \begin{minipage}{.3\textwidth}
  \centering
  \includegraphics[width=\linewidth]{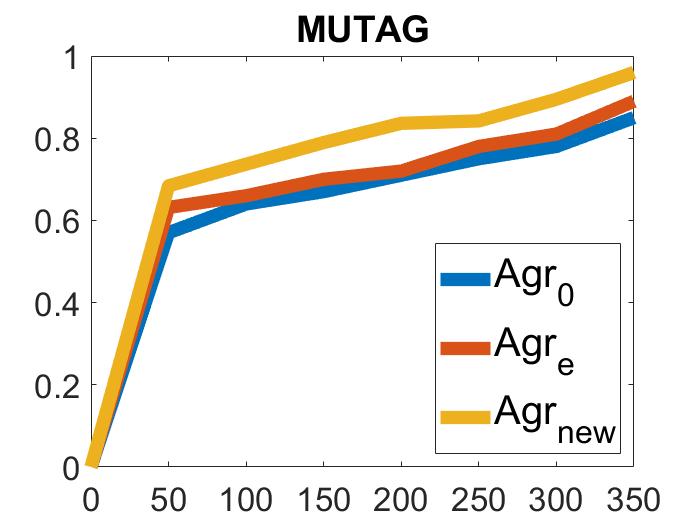}
  \end{minipage}
  \hfill
  \begin{minipage}{.3\textwidth}
  \centering
  \includegraphics[width=\linewidth]{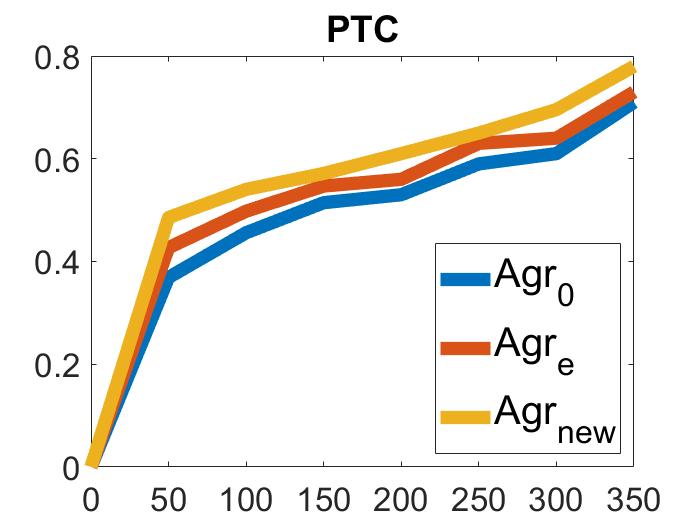}\end{minipage}
  \hfill
   \begin{minipage}{.3\textwidth}
  \centering
  \includegraphics[width=\linewidth]{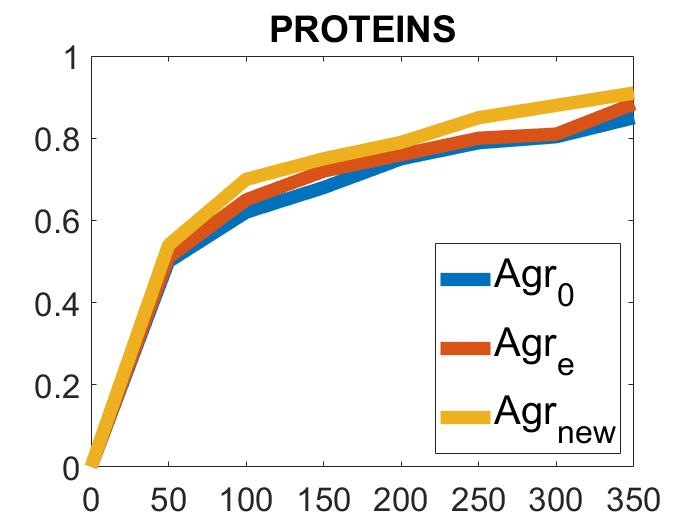}
  \end{minipage}
  \hfill
  \begin{minipage}{.3\textwidth}
  \centering
  \includegraphics[width=\linewidth]{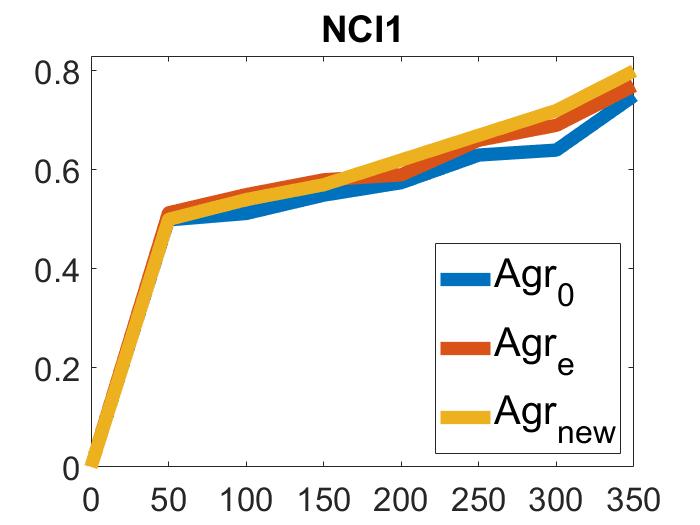}
  \end{minipage}
  \hfill
  \begin{minipage}{.3\textwidth}
  \centering
  \includegraphics[width=\linewidth]{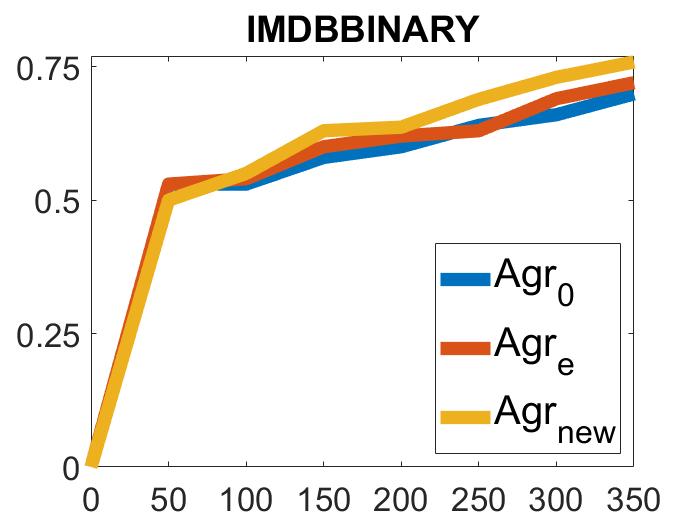}
  \end{minipage}
  \hfill
  \begin{minipage}{.3\textwidth}
  \centering
  \includegraphics[width=\linewidth]{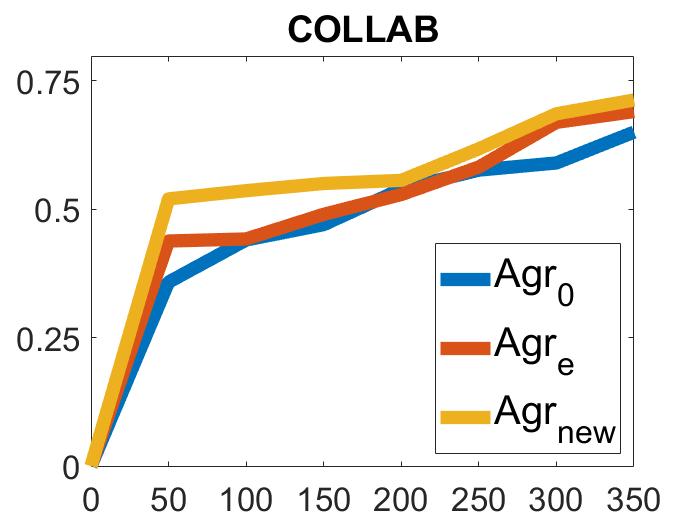}
  \end{minipage}
 \caption{Training set performance of IV-GNN and less powerful interval valued feature accepting GNNs. X-axis and Y-axis show the epoch number and the accuracy of training respectively.}
  \label{training}
\end{figure}

We have already theoretically analyzed the representational power of our proposed IV-GNN. Now, we validate our theoretical findings by comparing training accuracies on various datasets. Ideally, the architecture, which has stronger representational power, should fit the data more accurately, resulting in better training performance. Training set performance gives an idea about how well the model learned from training data. \\
Figure \ref{training} shows training curves of IV-GNN and interval-valued GNN with alternative aggregation functions with the same hyperparameter settings. In comparison, the GNN variants, which use less powerful interval aggregator as $AGGREGATE$ function, cannot learn from many datasets. The reason behind this observation is, $agr_{new}$ has more distinguishing power than $agr_{0}$ and $agr_{e}$. Between $agr_{0}$ and $agr_{e}$, $agr_{e}$ performs better because as we have seen theoretically that, with non repetitive feature value, $agr_{e}$ is equally powerful as $agr_{new}$.

\begin{figure}
  \begin{minipage}{.49\textwidth}
  \centering
  \includegraphics[width=\linewidth]{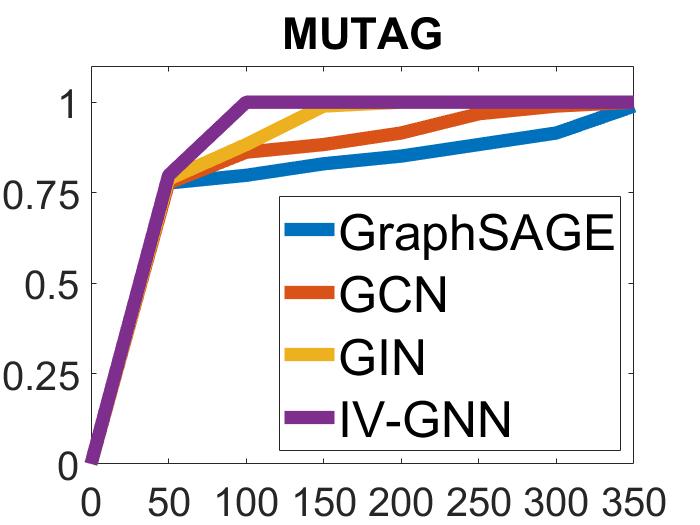}
  \end{minipage}
  \hfill
  \begin{minipage}{.49\textwidth}
  \centering
  \includegraphics[width=\linewidth]{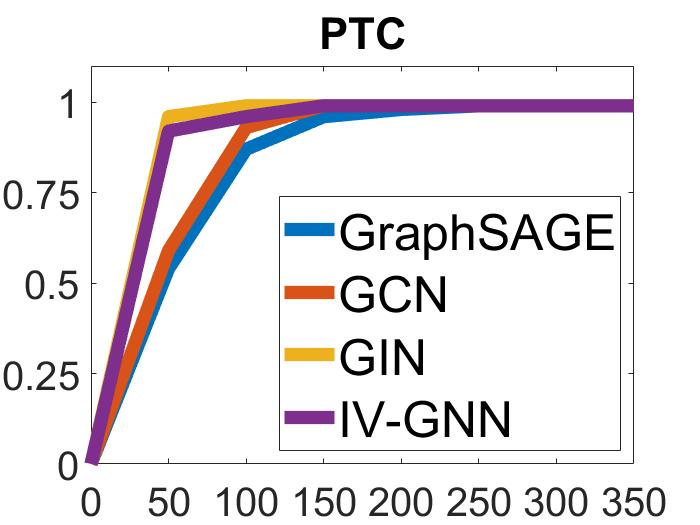}  
  \end{minipage}
  \newline
  \hfill
  \begin{minipage}{.49\textwidth}
  \centering
 \includegraphics[width=\linewidth]{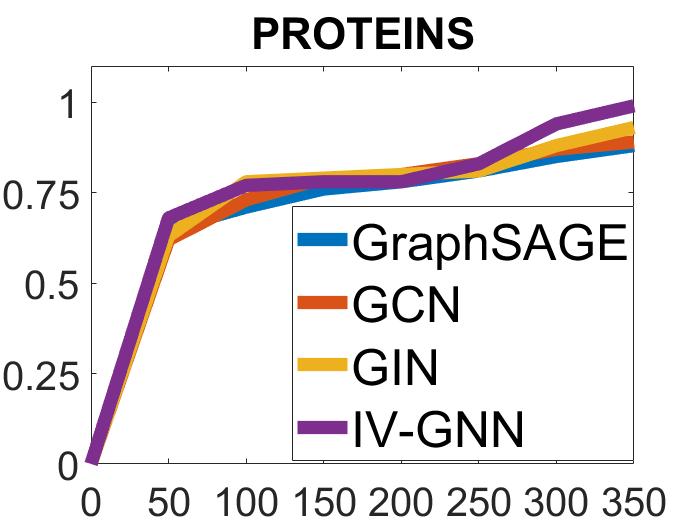}  
\end{minipage}
   \begin{minipage}{.49\textwidth}
   \centering
   \includegraphics[width=\linewidth]{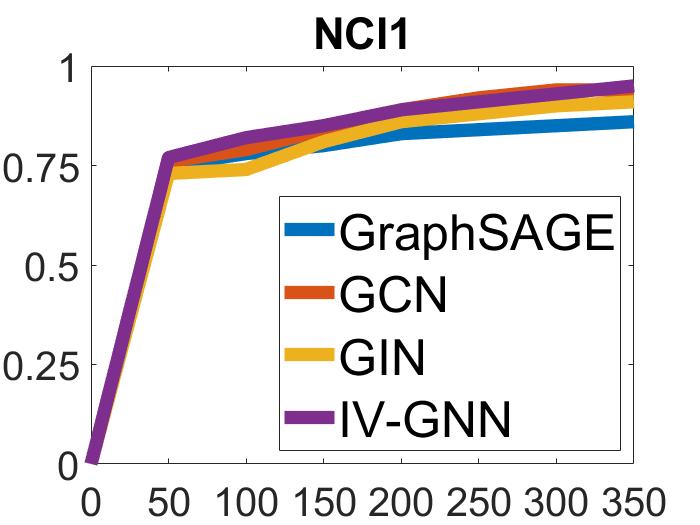}
  \end{minipage}
  \begin{minipage}{.49\textwidth}
   \centering
   \includegraphics[width=\linewidth]{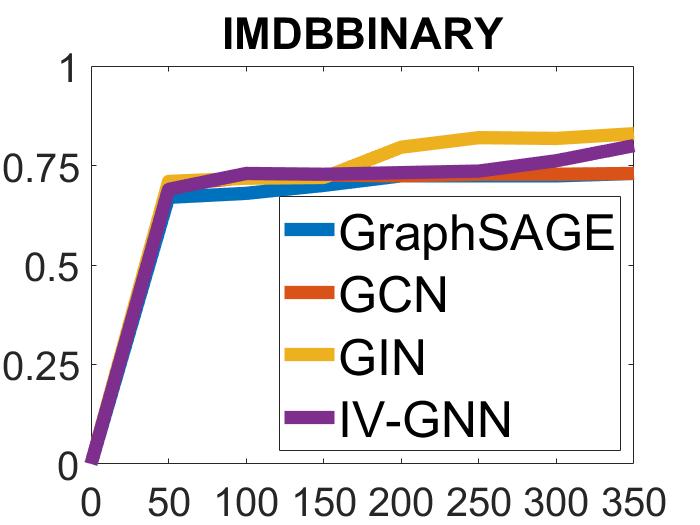}
  \end{minipage}
  \hfill
  \caption{Training set performance of IV-GNN accepting degenerate interval value and other countable value feature accepting GNN. X-axis and Y-axis show the epoch number and the accuracy of training respectively.}
  \label{figuredegenerate}
\end{figure}

Figure \ref{figuredegenerate} shows training curves of IV-GNN accepting degenerate interval valued feature and other state-of-the-art GNN models, who accept countable valued features. As we can see, curve of IV-GNN outgrows that of others, which proves that, IV-GNN learns from the data much better than other model in most of the cases. In cases of dataset IMDB-BINARY, GIN captures the dataset slightly better than IV-GNN . However, our proposed model IV-GNN is able to beat the other two models quite efficiently. 
\subsubsection{Test set performance}
Now, we compare test set accuracies with respect to different interval aggregation-based GNN model and state-of-the-art GNN model. 
\\
We have perform this experiment for the synthetic as well as real life datasets for $350$ epochs and report the best cross-validation accuracy mean and standard deviation averaged over the 10 folds after performing each experiment \emph{ten} times . In Table \ref{table1}, we have seen that IV-GNN can capture graph structures and generalize well in most of the cases among other variants showing $4\%$ better accuracy on an average.
\begin{table}[htbp]
\caption{Test set classification accuracies in percentage}
  \centering
  \resizebox{0.75\textwidth}{!}{\begin{minipage}{\textwidth}
        \begin{tabular}{|c|c|c|c|}
\hline
\textbf{Dataset} & $agr_{0}$\textbf{\textit{-based model}}& $agr_{e}$\textbf{\textit{-based model}}& $agr_{new}$\textbf{\textit{-based model IV-GNN}} \\
\hline
SYNTHETIC\_1\_200  & 89.33 $\pm$ 0.13 & 95.00 $\pm$ 0.19 & 98.39 $\pm$ 0.24 \\
\hline
SYNTHETIC\_1\_1000 & 87.82 $\pm$ 0.66 & 89.82 $\pm$ 0.42 & 95.01 $\pm$ 0.24 \\
\hline
SYNTHETIC\_1\_2000 & 81.80 $\pm$ 0.40 & 92.88 $\pm$ 0.36 & 96.05 $\pm$ 0.31 \\
\hline
SYNTHETIC\_2\_200 & 82.98 $\pm$ 0.10 & 92.48 $\pm$ 0.89 & 98.92 $\pm$ 0.39 \\
\hline
SYNTHETIC\_2\_1000  & 83.92 $\pm$ 0.55 & 
87.75 $\pm$ 0.26 & 92.10 $\pm$ 0.79 \\
\hline
SYNTHETIC\_2\_2000 & 
83.31$\pm$ 0.19 & 86.75 $\pm$ 0.27 & 88.54 $\pm$ 0.67 \\
\hline
MUTAG & 85.79 $\pm$ 0.04 & 89.85 $\pm$ 0.22 & 92.37 $\pm$ 0.26 \\
\hline
PTC & 61.10 $\pm$ 0.19 & 63.6 $\pm$ 0.23 & 67.6 $\pm$ 0.25 \\
\hline
PROTEINS & 78.09 $\pm$ 0.04 & 80.7 $\pm$ 0.54 & 83.3 $\pm$ 0.06 \\
\hline
NCI1 & 75.6 $\pm$ 0.05 & 77.7 $\pm$ 0.03 & 80.3 $\pm$ 0.10\\
\hline
IMDB-B & 69.24 $\pm$ 0.03 & 72.41 $\pm$ 0.12 & 72.5 $\pm$ 0.16\\
\hline
COLLAB & 65.6 $\pm$ 0.02 & 69.4 $\pm$ 0.02 & 71.1 $\pm$ 0.16\\
\hline
\end{tabular}
\end{minipage}}
\label{table1}      
\end{table}
\\
In the special case for degenerate intervals, we perform experiments \emph{ten} times on real-life datasets only and report the mean and standard deviation of the accuracies. We could not perform experiments of GraphSAGE and GCN on the dataset COLLAB due to memory bound. As we can see in the table \ref{table2}, IV-GNN performs much better than the other state-of-the-art approaches on four out of six datasets and achieves a performance gain of $7\%$ on an average. However, for the datasets like IMDB-BINARY and COLLAB, IV-GNN gives comparative results with GIN, whereas GraphSAGE and GCN fail to work on such large graph datasets.
\begin{table}[htbp]
\caption{Test set classification accuracies in percentage}
  \centering
  \resizebox{0.75\textwidth}{!}{\begin{minipage}{\textwidth}
        \begin{tabular}{|c|c|c|c|c|}
\hline
\textbf{Dataset} & \textbf{\textit{GraphSAGE}}& \textbf{\textit{GCN}}& \textbf{\textit{GIN}}& \textbf{\textit{IV-GNN}} \\
\hline
MUTAG & 85.4 $\pm$ 0.77 & 82.9 $\pm$ 0.66 & 89.4 $\pm$ 0.84 & 94.7 $\pm$ 0.24 \\
\hline
PTC & 63.3 $\pm$ 0.94 & 66.9 $\pm$ 0.19 & 64.6 $\pm$ 0.74 & 68.5 $\pm$ 0.34 \\
\hline
PROTEINS & 75.8 $\pm$ 0.34 & 76.23 $\pm$ 0.14 & 76.75 $\pm$ 0.98 & 88.1 $\pm$ 0.63  \\
\hline
NCI1 & 78.1 $\pm$ 0.34 & 79.2 $\pm$ 0.75 & 82.5 $\pm$ 0.11 & 83.92 $\pm$ 0.96 \\
\hline
IMDB-B & 71.38 $\pm$ 0.97 & 72.41 $\pm$ 0.9 & 74.1 $\pm$ 0.2 & 73.35 $\pm$ 0.68 \\
\hline
COLLAB & - & - & 80.2 $\pm$ 0.53 & 79.85 $\pm$ 0.3 \\
\hline
\end{tabular}
\end{minipage}}
\label{table2}      
\end{table}
\subsection{Empirical study on hyperparameter setting}\label{hyper}
\begin{figure}
  \begin{minipage}{.49\textwidth}
  \centering
  \includegraphics[width=\linewidth]{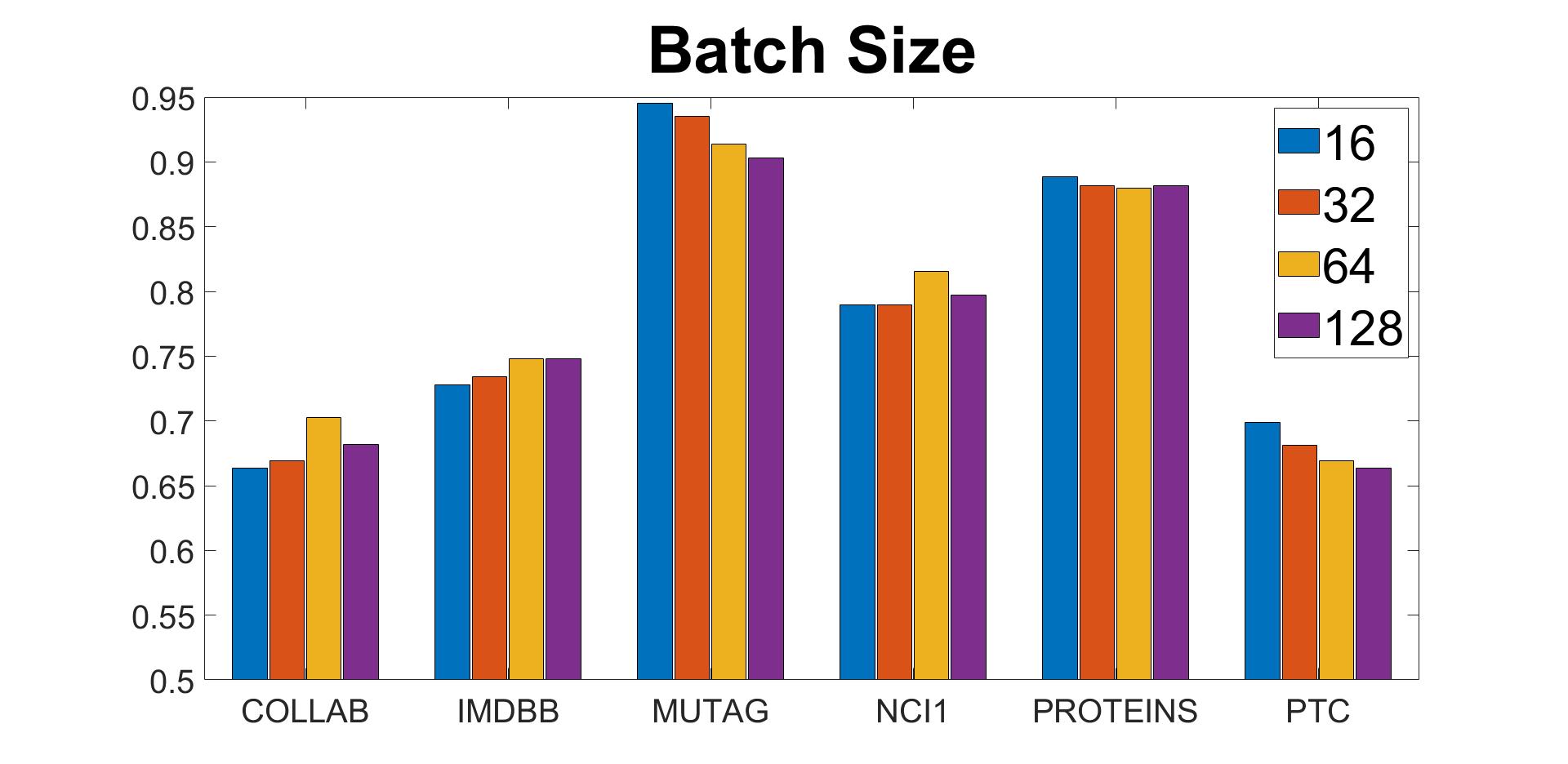}
  \end{minipage}
  \hfill
  \begin{minipage}{.49\textwidth}
  \centering
  \includegraphics[width=\linewidth]{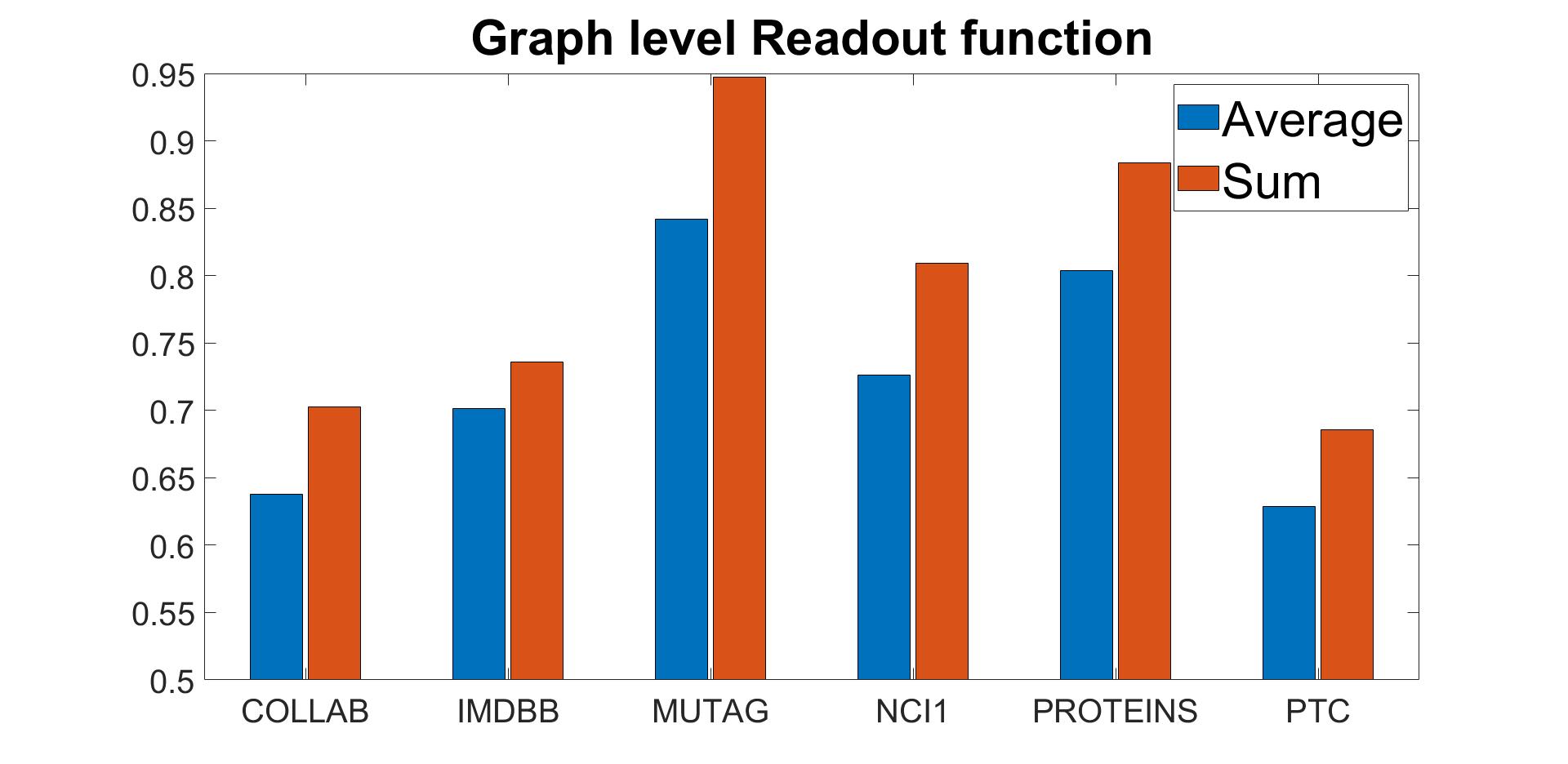}  
  \end{minipage}
  \newline
  \hfill
  \begin{minipage}{.49\textwidth}
  \centering
 \includegraphics[width=\linewidth]{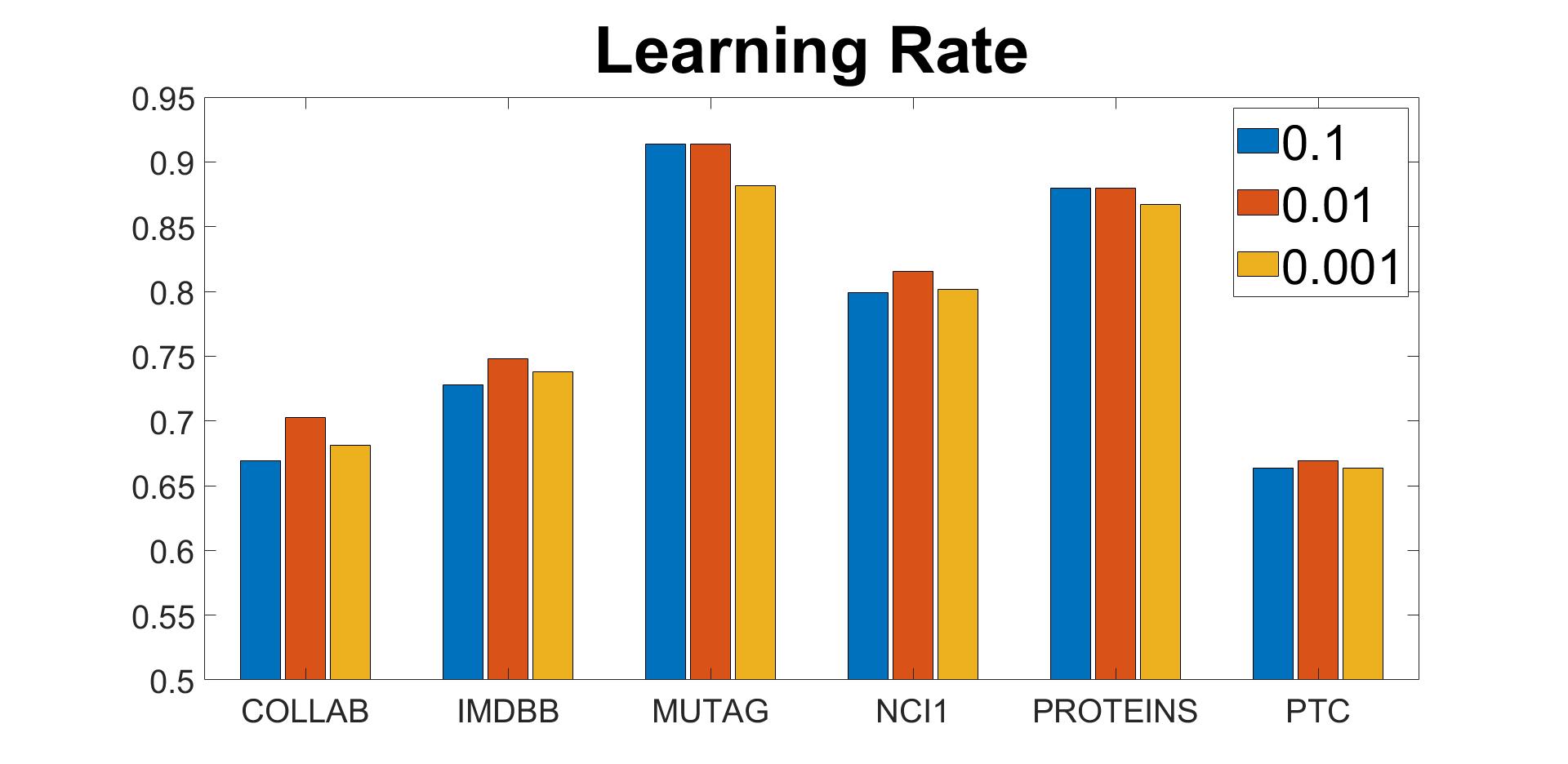}  
\end{minipage}
   \begin{minipage}{.49\textwidth}
   \centering
   \includegraphics[width=\linewidth]{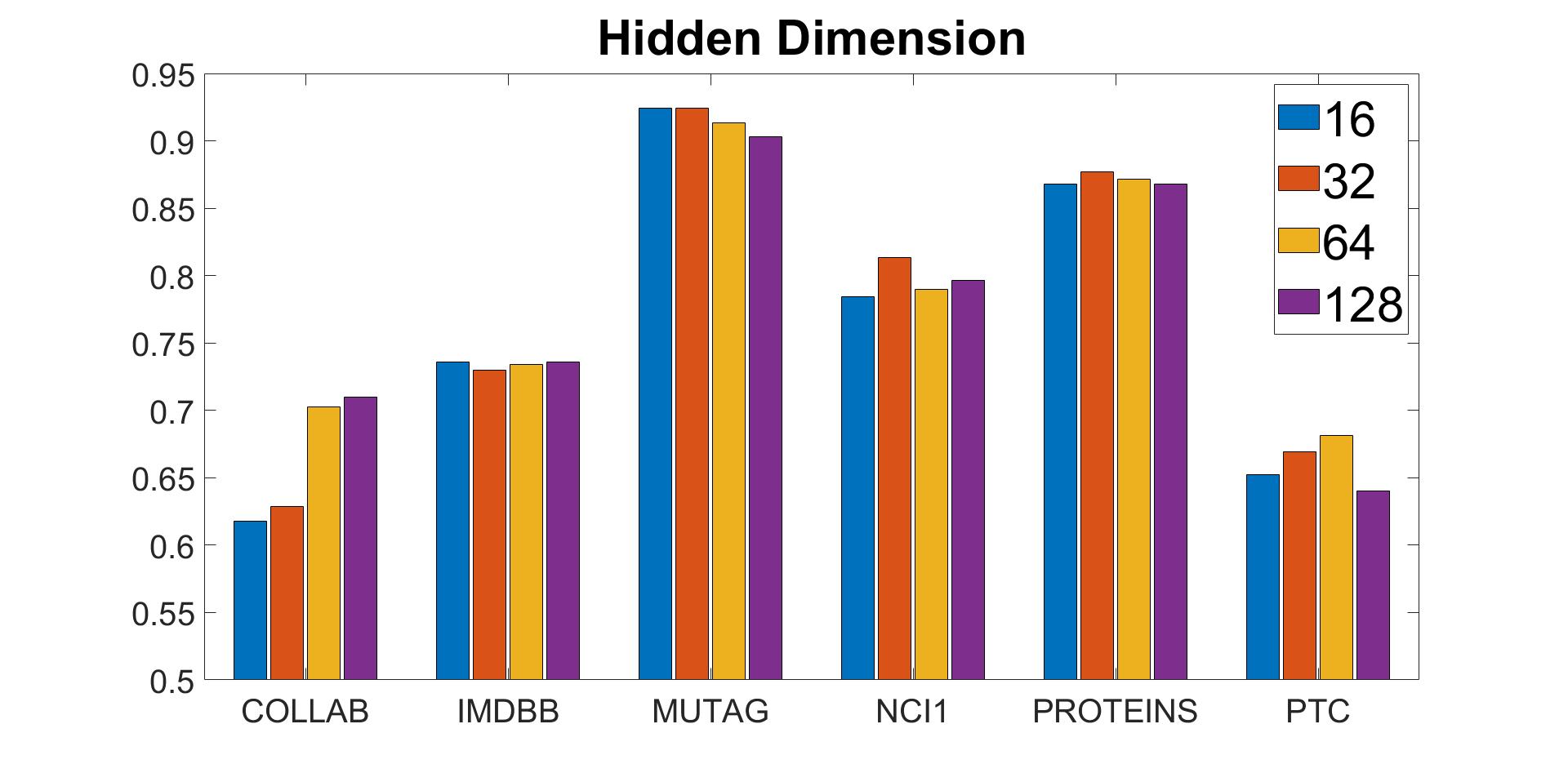}
  \end{minipage}
  \hfill
  \caption{empirical study on hyper-parameter setting. X-axis and Y-axis show the epoch number and the accuracy of training respectively.}
  \label{hyperpara}
\end{figure}

We have performed empirical study on various hyperparameters involved in the model and our experimental finding can be depicted in the figure \ref{hyperpara}.
\subsubsection{Effect of different Batch Size}
Batch size means the number of training examples to work through before updating the internal model parameters. We have experimented with the batch size $\{16,32,64,128\}$ and as we can see that for the relatively smaller dataset as MUTAG, PTC and PROTEINS, smaller batch size works better and to learn from the more enormous datasets such as COLLAB, NCI1 and IMDB-BINARY, larger batch size performs better. So we take batch size as $\{16,64\}$ depending on the size of the datasets.
\subsubsection{Effect of different Learning Rate}
Learning rate is a measure of the step size towards moving to the minimum of the loss function \cite{murphy2012machine}. A lower learning rate can slow down the convergence process, while a too high learning rate may jump over the minima and oscillates. Based on the experimental result, we find that between $\{0.1,0.01,0.001\}$, $0.01$ gives maximum accuracy for IV-GNN.
\subsubsection{Effect of different Hidden Dimension}
We have experimented with the different number of units in the hidden layer of the MLP, such as $\{16,32,64,128\}$. Here also, smaller hidden dimension works well for smaller datasets as larger hidden dimension may cause underfitting. So, we take $32$ as a hidden dimension for the bioinformatics dataset. On the other hand, for the social network datasets, hidden dimension $128$ captures the structures efficiently.  
\subsubsection{Effect of different Graph level Readout function}
We have experimented two different Graph level Readout functions $Sum$ and $Average$. As shown in the figure, $Sum$ performs much better than the function $Average$, which establishes the fact that $Sum$ have better representational capacity than $Average$.
\section{Conclusion}\label{conclu}
This paper developed a new interval aggregation scheme, having much better discriminative power than existing aggregation function. Using this newly developed aggregation operator as AGGREGATE function, we develop a Graph Neural Network based architecture IV-GNN, which relaxes the condition on the feature space of being countable. Despite of being much more general in nature, the proposed method far outperforms the state-of-the-art approaches on several synthetic and real-life datasets(comparable results for IMDB-BINARY and COLLAB).
Incidentally, the aggregation function, proposed in this work, is not continuous. Therefore, there may arise some situation where a slight change in aggregating intervals may bring  significant change in the resultant interval, which is not expected. In future, we would like to apply this architecture to perform different graph-related tasks such as node classification, link prediction etc. Also, there are architectures having summarizing ability of a continuous data such as LSTM. We would like to include the interval-valued feature as a summarized embedding and investigate quality of the performance on graph classification task. Another exciting direction for future work is exploring different interval aggregation according to the demand of the situation. 
\section*{Acknowledgement}
We want to acknowledge support from J.C.Bose Fellowship[SB/S1/JCB-033/2016 to S.B.] by the DST, Govt. of India. Also, we would like to thank Dr. Monidipa Das and Dr. Monalisa Pal for their valuable comments.

\nocite{*}
\bibliographystyle{fundam}
\bibliography{figuide}


\end{document}